\documentclass{article}

\usepackage{fullpage} 

\usepackage[utf8]{inputenc} 
\usepackage[T1]{fontenc}    
\usepackage{hyperref}       
\usepackage{url}            
\usepackage{booktabs}       
\usepackage{amsfonts}       
\usepackage{nicefrac}       
\usepackage{microtype}      

\usepackage{amsmath, amsthm, amssymb, mathtools, bbm}
\usepackage{algorithm, algpseudocode} 
\numberwithin{equation}{section}

\usepackage[usenames]{xcolor}
\usepackage{bm, xspace}
\usepackage{pifont}
\usepackage{multirow, hhline, array, graphicx, multicol}
\usepackage{tikz, pgfplots}

\usepackage{natbib}
\usepackage{makecell}
\usepackage{enumitem}

\newtheorem{Theorem}{Theorem}[section]
\newtheorem{Lemma}{Lemma}[section]
\newtheorem{Corollary}{Corollary}[section]
\newtheorem{Claim}{Claim}[section]

\newtheorem{Assumption}{Assumption}
\newtheorem{Remark}{Remark}[section]

\title{Benefits of Jointly Training Autoencoders: \\ An Improved Neural Tangent Kernel Analysis}

\author{Thanh V. Nguyen, Raymond K. W. Wong, Chinmay Hegde
   \thanks{TN is with the ECE Department at Iowa State University. RW is with the Statistics Department at Texas A\&M University. CH is with the Tandon School of Engineering at New York University. This work was supported in part by the National Science Foundation under grants DMS-1612985/1806063, CCF-1934904, CCF-1815101, CAREER CCF-2005804, and a faculty fellowship from the Black and Veatch Foundation.}
}

\begin{document}

\maketitle


\def\1{\mathbbm{1}}
\def\Id{\mathbb{I}}

\newcommand*{\bigcdot}{\bullet}

\newcommand\numberthis{\addtocounter{equation}{1}\tag{\theequation}}

\newcommand{\cA}[1]{A_{#1}} 
\newcommand{\cAi}{\cA{i}}
\newcommand{\cAj}{\cA{j}}
\newcommand{\cAl}{\cA{l}}
\newcommand{\cAS}{\cA{S}}

\newcommand{\cB}[1]{A_{#1}} 
\newcommand{\cBi}{\cB{i}}
\newcommand{\cBS}{\cB{S}}

\newcommand{\cD}[1]{D_{ #1}} 
\newcommand{\cDi}{\cD{i}}
\newcommand{\cDj}{\cD{j}}
\newcommand{\cDl}{\cD{l}}
\newcommand{\cDS}{\cD{S}}

\newcommand{\rA}[1]{A_{#1 \bigcdot}} 
\newcommand{\rAj}{\rA{j}}
\newcommand{\rAR}{\rA{R}}
\newcommand{\rAl}{\rA{l}}

\newcommand{\AR}[1]{A_{R, #1}}

\newcommand{\WT}{W^\top}
\newcommand{\WzrT}{W(0)^\top}
\newcommand{\wT}{w^\top}
\newcommand{\AT}{A^\top}
\newcommand{\aT}{a^\top}
\newcommand{\ZT}{Z^\top}
\newcommand{\zT}{z^\top}

\newcommand{\XT}{X^\top}
\newcommand{\xT}{x^\top}
\newcommand{\Xt}{\widetilde{X}}
\newcommand{\xt}{\widetilde{x}}
\newcommand{\Wt}{\widetilde{W}}
\newcommand{\wt}{\widetilde{w}}
\newcommand{\xtest}{x^{\mathrm{test}}}

\newcommand{\cg}[1]{g_{#1}} 
\newcommand{\cgi}{\cg{i}}
\newcommand{\cgS}{\cg{S}}
\newcommand{\gR}[1]{g_{R, #1}}
\newcommand{\Sc}{S^{\bot}}

\def\ghat{\widehat{g}}
\def\Rhat{\widehat{R}}
\def\dhat{\widehat{d}}
\def\ehat{\widehat{e}}
\def\Muv{M_{u,v}}
\def\Mhuv{\widehat{M}_{u,v}}
\def\flin{f_t^{\mathrm{lin}}}

\def\dsn{\mathfrak{D}} 
\def\event{\mathcal{E}} 
\def\fclass{\mathcal{F}} 
\def\normal{\mathcal{N}}
\def\Ne{N_{\varepsilon}}
\def\Rad{\mathcal{R}}
\def\Aupb{\mathcal{A}}
\def\Bupb{\mathcal{B}}
\def\di{\Delta_i}
\def\dj{\Delta_j}
\def\dl{\Delta_l}
\def\muij{\mu_{ij}}
\def\muil{\mu_{il}}
\def\mujl{\mu_{jl}}
\def\mux{\kappa_1}
\def\sigmax{\kappa_2}
\def\varx{\kappa}

\newcommand*{\Asmp}[1]{Assumption \textnormal{\textbf{A#1}}}
\newcommand*{\AsmpB}[1]{Assumption \textnormal{\textbf{B#1}}}

\def\sgn{\mathrm{sgn}}
\def\vec{\mathrm{vec}}
\def\supp{\mathrm{supp}}
\def\diag{\mathrm{diag}}
\def\card{\textnormal{card}}
\def\thres{\mathrm{threshold}}
\def\nor{\mathrm{normalize}}
\def\polylog{\textnormal{polylog}}
\def\iid{\text{i.i.d.}}
\def\whp{\text{w.h.p.}}
\def\wrt{\text{w.r.t.}}
\def\ie{\text{i.e.}}
\def\eg{\text{e.g.}}
\def\relu{\mathrm{ReLU}}
\def\df{\mathrm{d}}
\def\dt{\mathrm{d}t}
\def\der{\mathrm{d}}

\newcommand{\ceil}[1]{\left \lceil #1 \right \rceil}

\def\cmark{\ding{51}}
\def\xmark{\ding{55}}

\newcommand{\E}{\mathbb{E}}
\newcommand{\var}{\mathrm{Var}}
\newcommand{\Prob}{\mathbb{P}}
\newcommand{\sigmae}{\sigma_\eta}
\newcommand{\HardThres}{\mathcal{H}}

\newcommand{\R}{\mathbb{R}}

\DeclarePairedDelimiterX{\norm}[1]{\lVert}{\rVert}{#1} 
\DeclarePairedDelimiterX{\inprod}[2]{\langle}{\rangle}{#1, #2}
\DeclarePairedDelimiterX{\abs}[1]{\lvert}{\rvert}{#1}

\DeclarePairedDelimiterX{\bigO}[1]{(}{)}{#1}
\def\Otilde{\widetilde{O}}
\def\Omgtilde{\widetilde{\Omega}}

\newcommand{\sgnEvent}{\mathcal{F}_{x^*}}

\renewcommand{\qedsymbol}{$\blacksquare$}


\begin{abstract}

A remarkable recent discovery in machine learning has been that deep neural networks can achieve impressive performance (in terms of both lower training error and higher generalization capacity) in the regime where they are massively over-parameterized. Consequently, over the past year, the community has devoted growing interest in analyzing optimization and generalization properties of over-parameterized networks, and several breakthrough works have led to important theoretical progress. However, the majority of existing work only applies to supervised learning scenarios and hence are limited to settings such as classification and regression. 

In contrast, the role of over-parameterization in the \emph{unsupervised} setting has gained far less attention. In this paper, we study the inductive bias of gradient descent for two-layer over-parameterized autoencoders with ReLU activation. We first provide theoretical evidence for the memorization phenomena observed in recent work using the property that infinitely wide neural networks under gradient descent evolve as linear models. We also analyze the gradient dynamics of the autoencoders in the finite-width setting. We make very few assumptions about the given training dataset (other than mild non-degeneracy conditions). Starting from a randomly initialized autoencoder network, we rigorously prove the linear convergence of gradient descent in two learning regimes, namely: 
\begin{enumerate}
\item the \emph{weakly-trained} regime where only the encoder is trained, and 
\item the \emph{jointly-trained} regime where both the encoder and the decoder are trained. 
\end{enumerate}
Our results indicate the considerable benefits of joint training over weak training for finding global optima, achieving a dramatic decrease in the required level of over-parameterization. 

We also analyze the case of weight-tied autoencoders (which is a commonly used architectural choice in practical settings) and prove that in the over-parameterized setting, training such networks from randomly initialized points leads to certain unexpected degeneracies.
\end{abstract}
\section{Introduction}
\label{sec:intro}

Deep neural networks have achieved great success in a variety of applications such as image and speech recognition, natural language processing, and gaming AI. Remarkably,  neural networks that achieve the state-of-the-art performance
in each of these tasks are all massively over-parameterized, with far more weight parameters than the sample size of training data or the input dimension. Such networks can gain impressive performance in terms of both (near) zero training error and high generalization capacity, 
which seemingly contradicts the conventional wisdom of
bias-variance tradeoffs. 
Surprising enough is the fact that (stochastic) gradient descent or its variants can effectively find global and generalizable solutions.
Explaining this phenomenon has arguably become one of the fundamental tasks for demystifying deep learning.

As a consequence, there has been growing interest in 
understanding the power of the gradient descent for over-parameterized networks. Over the past year, a specific line of research \citep{li2018learning, allen2018convergence, zou2018stochastic, du2018_gradient, oymak2019moderate, arora2019fine, zou2019improved} has led to exciting theoretical progress.  In particular,
the seminal work of \citet{du2018_gradient}
shows that gradient descent on two-layer neural networks with $\relu$ activation provably converges to some global minimum at a geometric rate, provided a sufficiently large number of neurons that is of polynomial order in the sample size.
The key idea that leads to this result is the following: once the network is sufficiently wide, gradient descent does not change the individual weights much, but results in a non-negligible change in the network output that exponentially reduces the training loss with iteration count. This line of thinking has been subsequently refined and linked to the stability of a special kernel, called the \emph{neural tangent kernel} (NTK) \citep{jacot2018ntk}. \citet{arora2019fine} showed that the minimum eigenvalue of the limiting kernel governs both the algorithmic convergence and the generalization performance. 

Despite these exciting results, the majority of existing work has focused on
\emph{supervised} settings and hence are limited to tasks such as classification
and regression. In contrast, the role of over-parameterization in the
\emph{unsupervised} setting (for tasks such as reconstruction, denoising, and
visualization) has gained much less attention. An early related example in unsupervised learning can be traced back to learning over-complete dictionaries with sparse codes \citep{olshausen97_sc}. Another example is the problem of learning mixtures of $k$ well-separated spherical Gaussians, where \cite{dasgupta2007_mg} showed that initializing with $O(k\log k)$ centers enables expectation-maximization to correctly recover the $k$ components. 

Interesting (but limited) progress has been made towards understanding over-parameterization for autoencoders, a popular class of unsupervised models
based on neural networks.
\cite{zhang2019identity} provided an extensive study of training highly over-parameterized autoencoders using a \emph{single} sample. They empirically showed that when learned by gradient descent,  autoencoders with different architectures can exhibit two inductive biases: memorization (i.e., learning the constant function) and generalization (i.e., learning the identity mapping) depending on the non-linearity and the network depth.  \citet{memorization_ae} showed that over-parameterized autoencoder learning is empirically  biased towards functions that concentrate around the training samples and hence exhibits memorization. \citet{buhai2019autoenc} empirically showed that over-parameterization benefits learning in recovering generative models with single-layer latent variables (including the sparse coding model). 

However, there has been a lack of theoretical evidence that supports these observations. \citet{zhang2019identity} were able to prove a result for a simple one-layer linear case while  \citet{memorization_ae} also proved the concentration of outputs near the training examples for a single-layer network under a data-restrictive setting. Moreover, none of the above papers have rigorously studied the training \emph{dynamics} of autoencoder models. The loss surface of autoencoder training was first characterized in \citep{tran17}. Subsequently, \citet{nguyen2019_ae} proved that under-parameterized (and suitably initialized) autoencoders performed (approximate) proper parameter learning in the regime of asymptotically many samples, building upon techniques in provable dictionary learning; cf.~\citep{arora15_neural,nguyen18_double}.

\paragraph{Our contributions.} In this paper, we provide the first rigorous analysis of inductive bias of gradient descent and gradient dynamics of over-parameterized, shallow (two-layer) autoencoders. To examine the inductive bias, we use an infinite-width approximation to derive the output reconstruction in terms its input. For the gradient dynamics, we study different training schemes and establish upper bounds on the level of over-parameterization under which (standard) gradient descent, starting from randomly initialized weights, can linearly converge to global optima provided the training dataset obeys some mild assumptions.  Our specific contributions are as follows:

\begin{enumerate}
\item First, we build upon the results by \citet{lee2019wide} to characterize the evolution of autoencoder output via linearization and infinite-width approximation. Then, we establish the inductive bias of infinite-width autoencoders trained with gradient descent and provide insights into the memorization phenomena. While our analysis is asymptotic with respect to the network width, empirical results in \citep{lee2019wide, zhang2019identity} strongly suggest that similar phenomena are exhibited at finite widths as well.
\item Next, we extend the results by \citet{du2018_gradient} to the setting of over-parameterized two-layer autoencoders. This involves developing a version of the NTK for multiple outputs, which can be done in a straightforward manner by lifting the kernel matrix of a single output into a higher-dimensional space via Kronecker products.
\item Next, we study the gradient dynamics of the \emph{weakly-trained}\footnote{This distinction of weak- vs.\ joint-training has been introduced in earlier work such as \cite{arora2019cntk}.} case where the training is done only over the weights in the encoder layer. We obtain a bound on the number of hidden neurons (i.e., level of over-parameterization) required to achieve linear convergence of gradient descent, starting from random initialization, to global optimality.
\item Next, we study the gradient dynamics of the \emph{jointly-trained} case where 
both the encoder and decoder are trained with gradient descent. We obtain a bound analogous to the weakly-trained case for the level of over-parameterization required for global convergence.
Interestingly, our bound for over-parameterization in the jointly trained case is significantly better compared with the {weakly-trained} case.
\item Finally, we study a special family of autoencoders for which 
the encoder and decoder are \emph{weight-tied}, i.e., the two layers share the same weights (this is a common architectural choice in practical applications).
  For the weight-tied case, we show that even without any training, $O(d/\epsilon)$ hidden units are able to achieve $\epsilon$-test error where $d$ is the input dimension.
  Indeed, as the number of hidden unit increases, the autoencoder approximately recovers an identity map.
  Since the identity map is not particularly useful in representation learning, we speculate that training of weight-tied autoencoders under over-parameterization may lead to unexpected degeneracies.
\end{enumerate}

\paragraph{Techniques.} Our analysis extends the techniques of \citet{lee2019wide} and \citet{du2018_gradient} for analyzing the global convergence of gradient descent in overparameterized neural networks using the neural tangent kernel. The special case of autoencoder networks is somewhat more complicated since we now have to deal with multiple outputs, but the use of Kronecker products enables us to derive concise NTK's for our setting. 

The work of \citet{du2018_gradient} and subsequent papers study the weakly-trained case for the supervised setting where the second layer is fixed. We derive analogous bounds for the autoencoder setting. Moreover, we derive a new result for the jointly-trained case and obtain a significantly improved bound on the requisite level of over-parameterization. Our result is based on three key insights: 

\begin{enumerate}[label=(\roman*)]
\item the linearization enables us to derive the autoencoder's reconstruction for a given input as a linear combination of the training samples weighted by kernel scores;
\item thanks to the linear decoder, the corresponding kernel is \emph{smooth}, and the improved smoothness allows gradient descent to move greater amount from the initial point; and 
\item with this improved smoothness, we can derive a sharper characterization of the descent trajectory length in Frobenius norm instead of column-wise Euclidean norm. 
\end{enumerate}



\section{Overview of main results}

\textbf{Notation.} We use uppercase letters to denote matrices, 
and lowercase for vectors or scalars.
An expectation is the notation $C$ which represents a generic scalar constant, whose value can change from line to line.
A vector is interpreted as a column vector by default. We denote by $x_i \in \R^d$ the $i^{\textrm{th}}$-column (or sample) of the data matrix $X$, and $W = [w_1, \dots, w_m] \in \R^{d\times m}$ denotes a weight matrix.
Whenever necessary, we distinguish between the weight vector $w_r$ at different algorithmic  steps using an explicit $w_r(t)$ indexed by the step $t$. For a matrix $A = [a_1, \dots, a_m] \in  \R^{d\times m}$, $\vec(A) = [a_{11}, \dots, a_{d1}, \dots, a_{1m}, \dots a_{dm}]^\top$ vectorizes the matrix $A$ by stacking its columns. The symbol $\otimes$  denotes the Kronecker product.

We use $\mathcal{N}(\cdot)$ and $\mathrm{Unif}(\cdot)$ to denote the Gaussian and uniform distributions respectively. We simply write $\E_{w}$ instead of $\E_{w \sim \normal(0, I)}$ for brevity. Throughout the paper, we refer to an arbitrary $\delta \in (0, 1)$ as the failure probability of some event under consideration.

\subsection{Two-layer autoencoders}
Our goal is to understand the inductive bias and the learning dynamics of learning two-layer autoencoders with gradient descent. We focus on the two-layer autoencoder architecture with the rectified linear unit ($\relu$), defined by $\phi(z)=\max(z,0)$ for any $z\in\R$.
In below, when $\phi$ is applied to a vector or a matrix, the $\relu$ function is applied element-wisely. Given an input sample $x \in \R^d$, the autoencoder
returns a reconstruction $u\in\R^d$ of $x$, given by
\[
  u = \frac{1}{\sqrt{md}}A\phi(W^\top x) = \frac{1}{\sqrt{md}}\sum_{r=1}^m a_r\phi(\wT_r x) ,
\]
where 
$W = [w_1, \dots, w_m]$ and $A= [a_1, \dots, a_m]$ are weight matrices of the first (encoder) and second (decoder) layers respectively. We do not consider bias terms in this work. However, in principle, the bias vector for the hidden layer can be regarded as the last column of $W$ with the last dimension of $x$ always being 1.

\begin{Remark}[Choice of scaling factor]
  \label{rmkScalingFactor1}
  \normalfont Notice that we have scaled the output with $1/\sqrt{md}$, where $1/\sqrt{m}$ is the factor for the first layer and $1/\sqrt{d}$ for the second layer. Such scaling has been utilized in mathematical analyses of supervised networks \citep{jacot2018ntk} as well as of autoencoders \citep{li2018randomae}. Since the ReLU is homogeneous to scaling, such factors can technically be absorbed into the corresponding weight matrices $W$ and  $A$, 
but we find that keeping such factors explicit is crucial to understand the asymptotic behavior of neural network training as the network widths (i.e., $m$ in this case) go to infinity.
\end{Remark}

Let us now set up the problem. Suppose that we are given $n$ training samples $X = [x_1, x_2, \dots, x_n]$. We assume that each weight is randomly and independently initialized. Then, we train the autoencoder via gradient descent over the usual squared-error reconstruction loss:
\begin{equation}
\label{eqnGenEmpLoss}
  L(W, A) = \frac{1}{2}\sum_{i=1}^n\norm{x_i - \frac{1}{\sqrt{md}}A\phi(\WT x_i)}^2 = \frac{1}{2}\sum_{i=1}^n\norm{x_i - u_i}^2.
\end{equation}

Throughout the paper, unless otherwise specified, we make the following assumptions:

\begin{Assumption}
\label{astUnitNorm}
  All training samples are normalized, i.e., $\|x_i\|=1$ for $i=1,\dots, n$.
\end{Assumption}

We gather the training samples into the data matrix $X = [x_1, x_2, \dots, x_n]$ and define $\lambda_n \triangleq \|X^\top X\|$. 
Assumption~\ref{astUnitNorm} implies that $\norm{X}_F = \sqrt{n}$ and hence $1 \leq \lambda_n \leq n$. We regard $\lambda_n$ as a parameter that depends on the data geometry. For certain families of matrices (e.g., those with independent Gaussian entries), $\lambda_n \sim O(\max(n/d, 1))$, which can be $o(n)$ depending on how large $n$ is in terms of $d$.
We note that throughout our analysis, $X$ is regarded as fixed, 
and we will focus on the randomness in the weights.

\begin{Assumption}
\label{astMinEigen}
Consider a random vector $w \sim \normal(0, I)$ and define $\xt_i = \1[\wT x_i \ge 0]x_i$ for each $i \in [n]$. Let $\Xt = \bigl[\xt_1, \dots, \xt_n\bigr]$. Assume $\min(\lambda_{\min}(\E_{w}[\widetilde{X}^\top\widetilde{X}]), \lambda_{\min}(\E_{w}[\phi(X^\top w)\phi(\wT X)])) = \lambda_0 > 0$.
\end{Assumption}
The matrix $\E_{w}[\widetilde{X}^\top\widetilde{X}]$ is the so-called Gram matrix from the kernel induced by the $\relu$ transformation and has been extensively studied in \citep{xie2016diverse, tsuchida2017invariance, du2018_gradient, arora2019fine}. 
  Although this condition is difficult to interpret, one sufficient condition established in \citep{oymak2019moderate} (Lemma H.1 and Lemma H.2) is that as long as the squared minimum singular value $\sigma^2_{\min}(X \star X) > 0$ where $\star$ denotes the Khatri-Rao product, then Assumption \ref{astMinEigen} holds. In this sense, our assumption is similar to that of \citet{oymak2019moderate} and slightly weaker than of \citet{du2018_gradient}, which only require $\lambda_{\min}(\E_{w}[\widetilde{X}^\top\widetilde{X}]) > 0$.

The above assumptions about the data are relatively mild, which are in sharp contrast with assuming a specific generative model for the data (e.g., dictionary models, mixture of Gaussians \citep{nguyen2019_ae, buhai2019autoenc}) that have so far been employed to analyze autoencoder gradient dynamics.

\subsection{Learning dynamics}
Depending on which weight variables are being optimized, we consider three training regimes:

\begin{itemize}
\item \textbf{Weakly-trained case:}
  This corresponds to the regime where the loss function \eqref{eqnGenEmpLoss} is optimized over the weights $W$ while keeping $A$ fixed. A different form of weak training is to fix 
  the encoder and optimize \eqref{eqnGenEmpLoss} over $A$. Indeed, this practice is perhaps a folklore: it corresponds to standard kernel regression where the global convergence depends on the Hessian associated with random $\relu$ features. We do not pursue this case any further since kernel methods are well understood, but note in passing that the Hessian will eventually show up in our analysis.

\item \textbf{Jointly-trained case:} This corresponds to the regime that \eqref{eqnGenEmpLoss} is optimized over both $W$ and $A$.  This case matches practical neural network training, and performs better than the weakly trained case. We will show that the contrast between {weakly-trained} and {jointly-trained} cases arises due to the nature of the different NTK's and our analysis may pave the way to better understanding of autoencoder training. 

\item \textbf{Weight-tied case:} Weight-tying is another common practice in training autoencoders. Here, one sets the encoder and decoder weights to be the same, i.e., $A = W$, and optimizes \eqref{eqnGenEmpLoss} over the common variables $W$. We study this problem from the perspective of over-parameterization and show that this case leads to somewhat unexpected degeneracies.
\end{itemize}

We adopt the framework introduced in \citep{du2018_gradient}. Our proofs proceed generally as follows:

\begin{enumerate}
  \item[(i)] We will consider the continuous flow of the autoencoder outputs $U(t) = [u_{1}(t), u_{2}(t), \dots, u_{n}(t)] \in \R^{d\times n}$ corresponding to the samples in $X$ at time $t$. This continuous flow can be morally viewed as the execution of gradient descent with infinitesimal learning rate. This enables us to write:
\begin{align*}
  \frac{\df\vec(U(t))}{\dt}  &= 
                               K(t)\vec(X - U(t)),
\end{align*}
where $K(t)$ is a kernel matrix.

\item[(ii)] From this characterization, we can infer that the spectrum of $K(t)$ governs the dynamics of the outputs. To derive explicit convergence bounds, we will first prove that $K(0)$ has positive minimum eigenvalue with high probability. This is achieved via using concentration arguments over the random initialization. Then, we will upper-bound the movement of each individual weight vector from the initial guess and hence bound the deviation of $K(t)$ from $K(0)$ in terms of spectral norm.

\item[(iii)] By discretizing the continuous-time analysis, we will obtain analogous bounds for gradient descent with a properly chosen step size and show that gradient descent linearly converges to a global solution.
\end{enumerate}

Our convergence results are informally stated in the following theorems:

\begin{Theorem}[Informal version of Theorems \ref{thmGradFlow_Case1} and \ref{thmGradDescent_Case1}]
  \label{thmInformalWeak}
  Consider an autoencoder that computes output $u = \frac{1}{\sqrt{md}}A\phi(W^\top x)$ where the weight vectors are initialized with independent vectors $w_{r} \sim \mathcal{N}(0, I)$ and $a_{r} \sim \mathrm{Unif}(\{-1, 1\}^d)$ for all $r \in [m]$. For any $
  \delta \in (0, 1)$ and $m \ge C \frac{n^5d^4\lambda_n}{\lambda_0^4\delta^3}$  for some large enough constant $C$, the gradient descent over $W$ linearly converges to a global minimizer with probability at least $1-\delta$ over the randomness in the initialization.
\end{Theorem}

\begin{Theorem}[Informal version of Theorems \ref{thmGradFlow_joint} and \ref{thmGradDescent_joint}]
  \label{thmInformalJoint}
  Consider an autoencoder that computes output $u = \frac{1}{\sqrt{md}}A\phi(W^\top x)$ where the weight vectors are initialized with independent vectors $w_{r} \sim \mathcal{N}(0, I)$ and $a_{r} \sim \mathrm{Unif}(\{- 1, 1\}^d)$ for all $r \in [m]$. For any $
  \delta \in (0, 1)$ and $m \ge C\frac{nd\lambda^3_n}{\lambda_0^4\delta^2} $ for some large enough constant $C$, the gradient descent jointly over $W$ and $A$ linearly converges to a global minimizer with probability at least $1-\delta$ over the randomness in the initialization.
\end{Theorem}

\paragraph{Comparisons with existing work.} We summarize the quantitative implications of our results in Table \ref{tblResults}. In this table, we compare with \citet{du2018_gradient, oymak2019moderate,zou2019improved} that achieve the best known bounds to our knowledge. 

We emphasize that the factor $d$ in our bounds arises due to the fact that our network produces high-dimensional outputs (dimension $d$ in the case of autoencoders) while the previous works have focused on scalar outputs. Note also that the input dimension $d$ is implicitly hidden in $\lambda_0$ and $\lambda_n$. 

For {weakly-trained} networks with a single output, we (slightly) improve the order of over-parameterization: $m = \Omega\left(\frac{n^5\lambda_n}{\lambda_0^4\delta^3} \right)$ over the previous bound $\Omega\left(\frac{n^6}{\lambda_0^4\delta^3} \right)$ in \citet[Theorem 3.2]{du2018_gradient} by explicitly exposing the role of the spectral norm $\lambda_n$ of the data. 

For the {jointly-trained} regime, we obtain a significantly improved bound over \citet[Theorem 3.3]{du2018_gradient}. Our result is consistent with \citet[Theorem 6.3]{oymak2019moderate}, but we have both layers jointly trained; the proof technique in \citet[Theorem 6.3]{oymak2019moderate} is different from ours (bounding Jacobian perturbations),
and does not seem to be easily extended to the jointly trained case.

Let us better understand the intuition behind the bounds in Table \ref{tblResults} in terms of the dimension $d$ and the sample size $n$. We emphasize that in the fairly typical regime of machine learning where $n \ge d$ and $\lambda_n \sim n/d$, the level of over-parameterization for the single output is moderate (of order $n^4/d^3$). Since autoencoders have an output dimension $d$, the factor-$d$ in the bounds is natural in the jointly-trained case by characterizing the trajectory length by Frobenius norm
. This is consistent with the result in \cite{zou2019improved}. Our bound is different from that in \citet{zou2019improved} in that we make assumption on the minimum eigenvalue $\lambda_0$ while they assume a lower bound on the sample separation $\Delta$. A direct universal comparison between the two bounds is difficult; however, \citet{oymak2019moderate} shows an upper bound $\lambda_0 \ge \Delta/100n^2$. Finally, we note that initializing $A$ with i.i.d. Rademacher entries keeps our analysis in line with previous work, and an extension to Gaussian random initialization of $A$ should be straightforward.

\begin{table}[h!]
  \centering
  \begin{tabular}{|c|c|c|c|}
    \hline
    Regime & Reference & Single output  & Multiple output \\
    \hline
    \hline
    \multirow{3}{*}{Weakly-trained} & \cite{du2018_gradient} & $C\frac{n^6}{\lambda_0^4\delta^3}$ & \xmark \\
    \hhline{~---}
           & This work & $C\frac{n^5\lambda_n}{\lambda_0^4\delta^3}$ & $C\frac{n^5d^4\lambda_n}{\lambda_0^4\delta^3}$ \\
    \hhline{~---}
    & \cite{oymak2019moderate} & $C\frac{n\lambda_n^3}{\lambda_0^4}$ & \xmark \\
    \hline
    \multirow{3}{*}{Joint-trained} & \cite{du2018_gradient} & $C\frac{n^6\log(m/\delta)}{\lambda_0^4\delta^3}$ & \xmark \\
    \hhline{~---}
    & \cite{zou2019improved} & $C\frac{n^8}{\Delta^4}$ & $C\frac{n^8d}{\Delta^4}$ \\
    \hhline{~---}
    & This work & $C\frac{n\lambda_n^3}{\lambda_0^4\delta^2}$ & $C\frac{nd\lambda^3_n}{\lambda_0^4\delta^2}$ \\
    \hline
  \end{tabular}
  \vskip .1in
  \caption{\small \sl Comparison of our over-parameterization bounds with the known results in \citep[Theorem 3.2 and Theorem 3.3]{du2018_gradient}, \citep[Theorem 6.3]{oymak2019moderate} and \citep[Table 1]{zou2019improved}. Here, $d$ is the input dimension, $n$ is the training size, $\lambda_0$ is the smallest eigenvalue of the Gram matrix, $\lambda_n$ is the maximum eigenvalue of the covariance matrix and $C$ is some sufficiently large constant. $\Delta$ is the smallest distance between any pair of distinct training points. 
  }
  \label{tblResults}
\end{table}

\subsection{Inductive bias}
The following theorem establishes a result on the inductive bias of
the infinitely wide autoencoders trained with gradient descent.
\begin{Theorem}
  \label{thmInformalBias}
  Let $K^{\infty} = \E_{W(0), A(0)}[K(0)]$. Assume $\lambda_{\min}(K^{\infty}) > 0$ and let $\eta_{\mathrm{critical}} = 2(\lambda_{\max}(K^{\infty}) + \lambda_{\min}(K^{\infty}))^{-1}$. Under gradient descent with learning rate $\eta < \eta_{\mathrm{critical}}$, for every normalized $x \in \R^d$ as the width $m \rightarrow \infty$, the autoencoder output $f_t(x)$ at step $t$ converges to $\mu_t(x) + \gamma_t(x)$, with:
  \begin{align*}
    \mu_t(x) &\rightarrow \sum_{i=1}^n \Lambda_i x_i , \\
  \gamma_t(x) &\rightarrow f_0(x) -  \sum_{i=1}^n \Lambda_i f_0(x_i)
  \end{align*}
  where each $\Lambda_i \in \R^{d\times d}$ depends on the kernel score between the input $x$ and each training sample $x_i$ and $K^{\infty}$. $f_0(x)$ is the autoencoder reconstruction of $x$ at initialization.
\end{Theorem}
We prove this result in Section \ref{sec:asymptotic}. Essentially, Theorem \ref{thmInformalBias} generalizes the simple result in \citet[][Theorem  1]{zhang2019identity} to non-linear autoencoders and
multiple-sample training despite its asymptotic nature. The closer the new test input $x$ is to the span of training data $X$, the more its reconstruction concentrates around these seen points. This coincides with the observation about ``memorization'' by \citet{memorization_ae}.


\section{The Neural Tangent Kernel and Linearized Autoencoders}
\label{sec:kernels}

\subsection{NTK for general autoencoders}

Let us first derive the  neural tangent kernels for general autoencoders (possibly deep and with more than 2 layers) with multiple outputs in a compact form. Given $n$ i.i.d samples $X = [x_1, x_2, \dots, x_n]$ and the autoencoder $f(\theta, x)$, we consider minimizing the squared-error reconstruction loss:

\begin{equation*}
  L(\theta) = \frac{1}{2}\sum_{i=1}^n\norm{x_i - 
    f(\theta, x_i)}^2 = \frac{1}{2}\sum_{i=1}^n\norm{x_i - u_i}^2,
\end{equation*}
where $\theta$ is a vector that stacks all the network parameters (e.g. $W$ and $A$) and $u_i = f(\theta, x_i) \in \R^d$ denotes the corresponding output for every $i = 1, 2, \ldots, m$. The evolution of gradient descent on $L(\theta)$ with an infinitesimally small learning rate is represented by the following ordinary differential equation (ODE):


\begin{equation}
\label{eqnGradODE}
\frac{\df\theta(t)}{\dt} = -\nabla_{\theta}L(\theta(t)).
\end{equation}

The {time-dependent} NTK for autoencoders can be characterized as follows: 
\begin{Lemma}
  \label{lmNTK_general}
  Denote by $U(t) = [u_{1}(t), u_{2}(t), \ldots, u_{n}(t)] \in \R^{d\times n}$ the corresponding outputs of all the samples in $X$, i.e., $u_i(t) = f(\theta(t), x_i)$. The dynamics of $U(t)$ is given by the ODE:
\begin{align*}
  \frac{\df\vec(U(t))}{\dt}  &= 
                               K(t)\vec(X - U(t)),
\end{align*}
where $K(t)$ is an $nd\times nd$ positive semi-definite kernel matrix whose $(i, j)$-th block of size $d\times d$ is:
\[
  \left( \frac{\partial}{\partial \theta}f(\theta, x_i) \right) \cdot \left( \frac{\partial}{\partial \theta}f(\theta, x_j) \right)^\top.
\]
\end{Lemma}
\proof
Note that in the supervised learning setting with a single output, the $(i,j)$-th block is a single scalar equal to the inner product of two gradients. We prove this using simple calculus. The gradient of the loss over the parameters $\theta$ is
\begin{align*}
  \nabla_{\theta}L(\theta)  &= -\sum_{i=1}^n \frac{\partial u_i^\top}{\partial \theta} (x_{i} - u_{i}), 
\end{align*}
where  $\partial u_i/\partial \theta$ denotes the Jacobian matrix of the output vector $u_i$ with respect to $\theta$. Combining with \eqref{eqnGradODE}, the continuous-time dynamics of the prediction for each sample $i \in [n]$ is specified as
\begin{align*}
  \frac{\df u_{i}}{\dt} &= 
                      \frac{\partial u_i}{\partial \theta} (-\nabla_{\theta}L(\theta)) \\
                      &= 
                        \sum_{j=1}^n \frac{\partial u_i}{\partial \theta} \frac{\partial u_j^\top}{\partial \theta}  (x_{j} - u_{j}).
\end{align*}
Vectorizing $\frac{dU(t)}{dt}$, we get
\begin{align*}
  \frac{\df\vec(U(t))}{\dt}   &= 
                                K(t)\vec(X - U(t)),
\end{align*}
where $K(t)$ (or $K$) is an $nd\times nd$ matrix whose $(i, j)$-block is of size $d\times d$:
\[
K_{i, j} =  \frac{\partial u_i}{\partial \theta} \frac{\partial u_j^\top}{\partial \theta} = 
\left( \frac{\partial}{\partial \theta}f(\theta, x_i) \right) \cdot \left( \frac{\partial}{\partial \theta}f(\theta, x_j) \right)^\top.
\]
One can easily verify that $K(t)$ is positive semi-definite.

\qedhere

If the parameters $\theta(0)$ are assumed to be stochastic, then the (deterministic) neural tangent kernel (NTK) is defined as:
\begin{equation}
  (K^\infty)_{i, j} = 
  \E_{\theta(0)}\left [\Bigl(\left. \frac{\partial}{\partial \theta}f(\theta, x_i)\right|_{\theta=\theta(0)} \Bigr) \cdot \Bigl(\left. \frac{\partial}{\partial \theta}f(\theta, x_j)\right|_{\theta=\theta(0)} \Bigr)^\top \right].
\end{equation}
Note that $K^\infty$ is time-independent.
If the network is randomly initialized and its width is allowed to grow infinitely large, $K(t)$ converges to $K^\infty$, and remains constant during training.
Our goal is to show that if the width is sufficiently large (not necessarily infinite), then $K(t) \approx K(0) \approx K^\infty$, and the gradient dynamics are governed by the spectrum of $K^\infty$.

\subsection{Linearized Autoencoders}
\label{sec:asymptotic}
While the NTK allows us to analyze the gradient dynamics of autoencoders, it does not provide a straightforward characterization of the reconstruction given any new input. This makes it difficult to reason about the inductive bias of the over-parameterization and gradient descent for autoencoders, which were empirically studied in \citep{zhang2019identity, memorization_ae}. Here, we theoretically justify these results by using linearization and infinite approximation based on the result of \citet{lee2019wide}.

For the autoencoder $f(\theta, x)$, we denote by $\theta(t)$ the parameter vector at time $t$ and by $\theta(0)$ its initial value. Let us simplify the notation by denoting $f_t(x) = f(\theta(t), x)$ and $f_t(X) =[f_t(x_1), \dots, f_t(x_n)]$. Recall the training objective:
\[
  L(\theta) = \frac{1}{2}\sum_{i=1}^n\norm{x_i - 
    f(\theta, x_i)}^2,
\]
and the gradient flow characterization of the training dynamics:
\begin{equation}
\frac{\df\theta(t)}{\dt} = - \nabla_{\theta}L(\theta(t)).
\end{equation}
Consider the following linearized autoencoder via the first order Taylor expansion of $f_t(x)$ around $\theta(0)$:
\begin{align*}
  \flin(x) \triangleq f_0(x) + \frac{\partial f_0(x)}{\partial \theta} \cdot \omega(t).
\end{align*}
Here, $\omega_t = \theta(t) - \theta(0)$ is the parameter movement from its initialization. The first term $f_0(x)$ or the initial reconstruction of $x$ remains unchanged during training over $\theta$ whereas the second term captures the dynamics with respect to the parameters, governed by:
\begin{align*}
  \dot{\omega}(t) &= -\sum_{i=1}^n \left( \frac{\partial f_0(x_i)}{\partial \theta} \right)^\top (x_i - \flin(x_i)), \numberthis \label{eqnLinearizedodODE1} \\ 
  \dot{f}_t^{\mathrm{lin}}(x) &= - \sum_{i=1}^n \frac{\partial f_0(x)}{\partial \theta} \left( \frac{\partial f_0(x_i)}{\partial \theta} \right)^\top (x_i - \flin(x_i)).
   \\  &= - K_0(x, X) \vec(X - \flin(X)). \numberthis \label{eqnLinearizedodODE2}
\end{align*}
where we denote
\begin{align*}
  \nabla_\theta f_0(X)^\top &\triangleq \left[\left( \frac{\partial f_0(x_1)}{\partial \theta} \right)^\top, \dots, \left( \frac{\partial f_0(x_n)}{\partial \theta} \right)^\top \right], \\
  K_0(x, X) &\triangleq \frac{\partial f_0(x)}{\partial \theta}  \nabla_\theta f_0(X)^\top \in \R^{d\times nd}, \\
\mathcal{K}_0 &\triangleq \nabla_\theta f_0(X) \nabla_\theta f_0(X)^\top \in  \R^{nd\times nd}.
\end{align*}
The last quantity is known as the neural tangent kernel matrix evaluated at $\theta(0)$, which is presented in the earlier section. Following from \citet{lee2019wide}, we have the closed form solutions for the ODEs in \eqref{eqnLinearizedodODE1} and \eqref{eqnLinearizedodODE2} as follows:
\begin{align}
  \label{eqnLinearizedODESol}
  \omega(t) &= -\nabla_\theta f_0(X)^\top \mathcal{K}_0^{-1}(I-e^{- \mathcal{K}_0 t}) \vec(X- f_0^{\mathrm{lin}}(X)), \\
  \vec(\flin(X)) &= (I-e^{-\mathcal{K}_0 t}) \vec(X) + e^{- \mathcal{K}_0 t} \vec(f_0(X)).
\end{align}
Moreover, given any new input $x$, the lineared output is $\flin(x) = \mu_t(x) + \gamma_t(x)$ where the signal and noise terms are given by
\begin{align}
  \mu_t(x) &= K_0(x, X)\mathcal{K}_0^{-1}(I-e^{- \mathcal{K}_0 t}) \vec(X), \label{eqnLinearizePredSig} \\
  \gamma_t(x) &= f_0(x) -  K_0(x, X)\mathcal{K}_0^{-1}(I-e^{- \mathcal{K}_0 t}) \vec(f_0(X)) \label{eqnLinearizePredNoise}.
\end{align}
These equations characterize the dynamics of reconstruction (up to scaling) for the linearized network. Now, we establish the connection between the infinitely wide autoencoder and its linearized version, and prove Theorem \ref{thmInformalBias}.

\proof[Proof of Theorem \ref{thmInformalBias}] We simply invoke Theorem 2.1 in \citep{lee2019wide} for the autoencoder case. Denote by $K^{\infty} = \E_{W(0), A(0)}[K(0)]$ the neural tangent kernel of the two-layer autoencoder. Assume $\lambda_{\min}(K^{\infty}) >0$ and let $\eta_{\mathrm{critical}} \triangleq 2(\lambda_{\max}(K^{\infty}) + \lambda_{\min}(K^{\infty}))^{-1}$. \citet{lee2019wide} shows that under gradient descent with learning rate $\eta < \eta_{\mathrm{critical}}$, for every $x \in \R^d$ such that $\norm{x} \le 1$, as the width $m \rightarrow \infty$, the autoencoder $f_t(x)$ converges in $\flin(x)$ given by Equation \eqref{eqnLinearizePredSig} and Equation \eqref{eqnLinearizePredNoise}.
\qedhere





\subsection{NTK for two-layer autoencoders}

Let us now specialize to the case of two-layer autoencoders with $\relu$ activation. Since we consider the two training regimes, including the \emph{weakly-trained} and \emph{jontly-trained}, 
we first give the expression of a few \emph{base} kernels whose appropriate compositions produce the final kernel for each individual case. The precise derivation of each regime is given in the next few sections.

Again, we consider the reconstruction loss:
\begin{equation*}
  L(W, A) = \frac{1}{2}\sum_{i=1}^n\norm{x_i - \frac{1}{\sqrt{md}}A\phi(\WT x_i)}^2 = \frac{1}{2}\sum_{i=1}^n\norm{x_i - u_i}^2,
\end{equation*}
where the weights are independently initialized such that:
\[
  w_{r}(0) \sim \mathcal{N}(0, I), ~~ a_{r}(0) \sim \mathrm{Unif}\{-1, 1\}^d,~~r = 1,\ldots, m .
\]
Here the minimization can be either over the encoder weights $W$, or the decoder weights $A$, or both $W$ and $A$. Let us denote
\[
\widetilde{X}_r(t) = \Bigl[\1[w_r(t)^\top x_{1} \ge 0]x_{1}, \ldots, \1[w_r(t)^\top x_{n} \ge 0]x_{n}\Bigr].
\] 
If we fix $A$ and optimize the loss $L(W, A)$ over $W$, we get
\[
  G(t) = \frac{1}{md} \sum_{r=1}^m\Xt_r(t)^\top\Xt_r(t) \otimes a_r\aT_r .
\] 
If we fix $W$ and optimize the loss $L(W, A)$ over $A$, we get 
\[
  H(t) = \frac{1}{md}\sum_{r=1}^m \phi(X^\top w_r(t))\phi(w_r(t)^\top X) \otimes I,
\]
Writing these kernels in Kronecker product form allows us to clearly visualize the connection to the supervised learning case, and enables characterization of their spectrum. Intuitively, in the {jointly-trained} case, since both $W$ and $A$ depend on $t$, an invocation of the chain rule leads to the sum $G(t) + H(t)$ being the ``effective'' kernel that governs the dynamics.


In the infinite-width limit where $m \rightarrow \infty$, the NTKs in the corresponding training regimes reduce to compositions of the following fixed deterministic kernels:
\begin{align*}
  G^{\infty} &= \E_{w(0), a(0)}\bigl[\widetilde{X}(0)^\top\widetilde{X}(0) \otimes a(0)a(0)^\top \bigr] = \E_{w(0)}[\widetilde{X}(0)^\top\widetilde{X}(0)] \otimes I_d,  \\
  H^{\infty} &= \E_{w(0)}\biggl[ \phi(X^\top w(0))\phi(w(0)^\top X) \biggr] \otimes I. 
\end{align*}


Somewhat curiously, we will show that the crucial component of the \emph{time-dependent} kernel in the jointly-trained regime, $H(t)$ (within $H(t)+G(t)$), is better-behaved than the corresponding kernel in the weakly-trained regime, $G(t)$, thanks to its better Lipschitz smoothness, even though the respective \emph{limiting} kernels are the same.
This improved smoothness allows us to derive a much better bound on kernel perturbations with respect to changing weights, and this results in a significant improvement in the level of over-parameterization (Theorem \ref{thmInformalJoint}).


\section{Inductive Biases of Over-parameterized Autoencoders}
In principle, the training dynamics of over-parameterized autoencoders are similar to those of supervised networks. However, the generalization properties or inductive biases of the over-parameterization are different and underexplored. In this section, we rigorously analyze the observations in \citep{zhang2019identity, memorization_ae} using the results we have developed.


\subsection{One-sample training}

This training setting was exclusively studied in \cite{zhang2019identity} with interesting insights on the memorization phenomemnon and the role of the depth and width. They were able to give some theoretical evidence for their observation in a simple one-layer linear case. Using linearization, we generalize this result for non-linear networks. We particularly focus on the two-layer architecture, but the results can be extended to networks of any depth. Although our result is asymptotic, \citet{lee2019wide} showed that networks with finite, large width exhibit the same inductive bias.

Suppose we have access to only one sample $x$ or the training data $X = x$. For a test input $x'$, $\flin(x') = \mu_t(x') + \gamma_t(x')$ where
\begin{align}
  \mu_t(x') &= K_0(x', x)\mathcal{K}_0^{-1}(I-e^{-\mathcal{K}_0 t}) x, \\
  \gamma_t(x') &= f_0(x') -  K_0(x', x)\mathcal{K}_0^{-1}(I-e^{- \mathcal{K}_0 t}) f_0(x).
\end{align}
As the learning rate for gradient desent is sufficiently small, the autoencoder output $f_t(x') \rightarrow \flin(x')$ as $m \rightarrow \infty$. In the infinite-width limit, the neural tangent kernel at $t=0$ converges to:
\begin{align*}
  \mathcal{K}_0 &= \frac{1}{md} \sum_{r=1}^m \1[w_r(0)^\top x \ge 0] a_r(0)a_r(0)^\top  + \phi(w_r(0)^\top x)^2 I \\
  &\rightarrow \frac{1}{d}\E[\1[w_r(0)^\top x \ge 0] a_r(0)a_r(0)^\top  + (w_r(0)^\top x)^2 I] = \frac{I}{d},
\end{align*}
since $w_r(0), a_r(0)$ are independent and $w_r(0)^\top x \sim \normal(0, 1)$. Therefore, 
the reconstruction is governed by the similarly between $x'$ and $x$ via the kernel function. Specifically,
\begin{align*}
  \mu_t(x') &\rightarrow d \cdot K_0(x', x)(1- e^{-t/d}) x, \\
  \gamma_t(x') &\rightarrow f_0(x') -  d \cdot K_0(x', x) (1- e^{- t/d}) f_0(x).
\end{align*}
Moreover, as $m \rightarrow \infty$, 
%
\begin{align*}
  K_0(x', x) &= \frac{\partial f_0(x')}{\partial \theta} \left( \frac{\partial f_0(x)}{\partial \theta} \right)^\top \\
             &= \frac{1}{md}\sum_{r=1}^m 1[w_r(0)^\top x' \ge 0, w_r(0)^\top x \ge 0] x'^\top x a_r(0)a_r(0)^\top + \phi(w_r(0)^\top x')\phi(w_r(0)^\top x) I \\
             &\rightarrow \frac{1}{d}\E_w[1[w^\top x' \ge 0, w^\top x \ge 0] x'^\top x] + \E_w[\phi(w^\top x')\phi(w^\top x) I] \\
  &= \langle x', x\rangle \frac{\pi - \arccos(\langle x', x\rangle)}{\pi d}\cdot I + \frac{1}{2\pi d}\sqrt{1 - \langle x', x\rangle^2} I.
\end{align*}
When $x'$ is close to $x$, $K_0(x', x) \sim I/d$, the signal term $\mu_t(x') \approx d \cdot K_0(x', x) x \approx x$ dominates the zero-mean noise term $\gamma_t(x') \approx f_0(x') -  f_0(x)$, then the reconstruction is close to $x$ that explains the memorization. When $x'$ is far from $x$, $\mu_t(x') \sim 0$ while $\gamma_t(x')$ is a random, so the reconstruction is governed by a random noise. See \citep{zhang2019identity} for more details on the empirical evidence.

\subsection{Multiple-sample training}
For the training with many samples, \citet{memorization_ae} showed that overparameterized autoencoders exhibit memorization by learning functions that
concentrate near the training examples. They proved that single-layer autoencoders project data onto the span of the training examples. We provide another intuition based on the reconstruction of the linearized networks. For an arbitrary input $x'$,
\begin{align*}
  \mu_t(x') &= K_0(x', X)\mathcal{K}_0^{-1}(I-e^{- \mathcal{K}_0 t}) \vec(X),\\
  \gamma_t(x') &= f_0(x') -  K_0(x', X)\mathcal{K}_0^{-1}(I-e^{- \mathcal{K}_0 t}) \vec(f_0(X)).
\end{align*}
The signal part of the reconstruction is a linear combination of training samples weighted by the kernel $K_0(x', x_i)$ and the eigenvalues of the kernel matrix $\mathcal{K}_0$. 
Therefore, as $m \rightarrow \infty$ and $t$ is sufficiently large,
\begin{align*}
  \mu_t(x') &\rightarrow \sum_{i=1}^n d\cdot K_0(x', x_i) x_i,\\
  \gamma_t(x) &\rightarrow f_0(x) - \sum_{i=1}^n d\cdot K_0(x', x_i) x_i f_0(x_i).
\end{align*}
The closer the new test input $x'$ is to the span of training data $X$, the more its reconstruction concentrates around these seen points. This coincides with the observation about ``memorization'' by \citet{memorization_ae}. 


\section{Weakly-trained Autoencoders}

We now analyze various training regimes; these will follow from different compositions of the above NTK's. In each of the analyses, we will first set up the corresponding NTK, study the gradient dynamics with infinitesimal step size (gradient flow), and then appropriately discretize the flow to get our final results.

\subsection{Gradient flow}
\label{sec:gradflow_case1}

Consider the {weakly-trained} regime with the objective function:
\begin{equation}
\label{eqnEmpLoss_Case1}
   L(W) = \frac{1}{2}\sum_{i=1}^n\norm{x_i - \frac{1}{\sqrt{md}}A\phi(W^\top x_i)}^2,
\end{equation}
where the corresponding minimization is \emph{only} performed over $W$. Suppose that the weight matrices $W$ and $A$ are randomly initialized such that
\[
  w_{ij}(0) \sim \mathcal{N}(0, 1), ~ a_{ij}(0) \sim \mathrm{Unif}(\{\pm 1\})
\]
are drawn independently for each all $(i, j)$. After the initialization, we keep $A$ fixed throughout and apply gradient descent learning over $W$ with step size $\eta$:
\[
  W(k+1) = W(k) - \eta \nabla_W L(W(k)),~k=0, 1, 2, \ldots .
\]
Let us derive the neural tangent kernel for this training regime. We first calculate the gradient of $L(W)$ with respect to $W$. Since $A\phi(\WT x) = \sum_{r=1}^ma_r\phi(w_r^\top x)$ for any $x\in\mathbb{R}^d$, it is convenient to compute the gradient with respect to each column $w_r$. The gradient $\nabla_{w_r}L(W)$ of the loss in~\eqref{eqnEmpLoss_Case1} over $w_r$ is given by:
\begin{align*}
  \nabla_{w_r}L &=  - \sum_{i=1}^n J_{r}(u_{i}) ^\top (x_{i} - u_{i})
                  = -\frac{1}{\sqrt{md}} \sum_{i=1}^n \1[w_r^Tx_i \ge 0]a_r\xT_i (x_{i} - u_{i}), \numberthis
                  \label{eqnGradLossOverWr_Case1}
\end{align*}
where  $J_{r}(u_i)$\footnote{Note that $\phi(z)$ is differentiable everywhere except at $z =0$, at which the derivative will be considered as 0.} denotes the Jacobian matrix of the output vector $u_i$ with respect to $w_r$:
\begin{align*}
  J_{r}(u_i) &= \frac{1}{\sqrt{md}}a_r\xT_i\phi'(\wT_rx) = \frac{1}{\sqrt{md}}\1[w_r^Tx_i \ge 0]a_r\xT_i. \numberthis
  \label{eqnJacobianUtowr_Case1}
\end{align*}

Let us consider the gradient flow for the weight vector $w_r(t)$ via the following ODE:
\begin{equation}
\label{eqnGradODE_Case1}
\frac{\df w_r(t)}{\dt} = -\nabla_{w_r}L(W(t)).
\end{equation}
Using~\eqref{eqnGradLossOverWr_Case1} and~\eqref{eqnGradODE_Case1}, the continuous-time dynamics of the prediction for each sample $i \in [n]$ is:
\begin{align*}
  \frac{\df u_{i}}{\dt} &= \sum_{j=1}^n \left( \sum_{r=1}^m J_{r} (u_{i}) J_{r}^\top(u_{j})\right) (x_{j} - u_{j}).
\end{align*}
Vectorizing $\frac{dU(t)}{dt}$, we get the equation that characterizes the dynamics of $U(t)$:
\begin{align*}
\label{eqnPredDynamics_Case1}
  \frac{\df \vec(U(t))}{\dt}
  &= \frac{1}{d}K(t)\vec(X - U(t)), \numberthis
\end{align*}
where $K(t)$ is the $nd\times nd$ matrix whose $(i, j)$-block is of size $d\times d$ and defined as
\[
  K(t)_{i, j} = d \sum_{r=1}^m J_{r} (u_{i}) J_{r}^\top(u_{j}) =  \frac{1}{m}\sum_{r=1}^m\1[w_r(t)^Tx_i \ge 0, w_r(t)^Tx_j \ge 0]x_{i}^\top x_{j}a_ra_r^\top.
\]
If we denote
\[
  \Xt_r(t) = \Bigl[\1[w_r(t)^\top x_{1} \ge 0]x_{1}, \dots, \1[w_r(t)^\top x_{n} \ge 0]x_{n}\Bigr].
\]
then we can write $K(t)$ in Kronecker form: 
\[
  K(t) = \frac{1}{m}\sum_{r=1}^m \Xt_r(t)^\top \Xt_r(t) \otimes a_r\aT_r.
\]
Since $W(0)$ and $A(0)$ are randomly initialized, in the limit as $m \rightarrow \infty$, $K(0)$ converges to the NTK:
\begin{align*}
  K^\infty &= \E_{W(0), A(0)}[K(0)] \\&= \E_{w(0), a(0)}[\Xt(0)^\top \Xt(0) \otimes a(0) a(0)^\top] \\&= \E_w[\Xt^\top \Xt] \otimes I,
\end{align*}
where the last step follows from the independence of $w(0)$ and $a(0)$.

By Assumption~\ref{astMinEigen}, $\lambda_{\min}(K^\infty) = \lambda_{\min}(\E_w[\Xt^\top \Xt]) = \lambda_0 >0$. In other words, the NTK kernel is strictly positive definite. We want to bound the minimum eigenvalue of $K(0)$ at the initialization $W(0)$ and prove $K(t) \approx K(0) \approx K^\infty$ when $m$ is large enough.

Now, we state the main theorem for the convergence of the gradient flow:
\begin{Theorem}[Linear convergence of gradient flow, weakly-trained regime]
  \label{thmGradFlow_Case1}
  Suppose Assumptions \ref{astUnitNorm} and \ref{astMinEigen} hold. Suppose at initialization that the weights are independently drawn such that $w_r \sim \normal(0, I)$ and $a_r \sim \mathrm{Unif}(\{\pm 1\}^d)$ for all $r \in [m]$. If $m \ge C\frac{n^5d^4\lambda_n}{\lambda_0^4\delta^2}$  for a constant $C > 0$, then with probability at least $1-\delta$
  \[
    \norm{X - U(t)}_F^2 \leq \exp\Bigl(-\frac{\lambda_0 t}{d}\Bigr)\norm{X - U(0)}_F^2.
  \]
\end{Theorem}
To prove this theorem, we use the auxiliary results from Lemmas \ref{lmMinEigenK0_Case1}, \ref{lmBoundKDiff_Case1} and \ref{lmMovementInTime_Case1}.

\begin{Lemma}
  \label{lmMinEigenK0_Case1}
  For any $\delta \in (0, 1)$, if $m \ge C\frac{\lambda_n^2d\log(nd/\delta)}{\lambda^2_0}$ for some large enough constant $C$,
  then with probability at least $1 -\delta$,
  one obtains $\norm{K(0) - K^\infty} \leq {\lambda_0}/{4}$ and $\lambda_{\min}(K(0)) \geq 3\lambda_0/4$ .
\end{Lemma}
The proof of this Lemma is given in Appendix \ref{appdx:weakly_train}.
\begin{Remark}
  \normalfont
  Compared with the results in~\citep{du2018_gradient, song2019quadratic}, our bound exposes the dependence on the data $X$ through the spectral norm of $X$ and the dimension $d$. When $\lambda_n$ is much smaller than $n$, our bound improves over these aforementioned results. For example, if the training samples are drawn from certain distributions (e.g., Gaussians, or from sparsely used dictionary models), the bound can be   as low as $m \sim \Otilde(d)$. 
\end{Remark}

The next step in our analysis is to upper bound the spectral norm of the kernel perturbation, $\norm{K(t) - K(0)}$, with high probability.

\begin{Lemma}
  Suppose $w_r \sim \normal(0, I)$ and $a_r \sim \mathrm{Unif}(\{\pm 1\}^d)$ are drawn independently for all $r \in [m]$. For any $\delta \in (0, 1)$ and some $R >0$, with probability at least $1-\delta$:
  \begin{equation}
    \label{eqn:suptwK}
    \sup_{\substack{\{\widetilde{w}=(\widetilde{w}_1,\dots,\widetilde{w}_m):\norm{\widetilde{w}_r - w_r} \leq R \\ ~\forall r \in [m]\}}} \norm{K(\widetilde{w}) - K(w)} < \frac{2n^2dR}{\delta},
  \end{equation}
  where $K(w) = \frac{1}{m}\sum_{r=1}^m\Xt(w_r)^\top \Xt(w_r) \otimes a_ra_r^\top$.
  \label{lmBoundKDiff_Case1}
\end{Lemma}

\begin{Remark}
  \normalfont
  One may ask why not to directly bound $K(t) - K(0)$ for each time $t$ but need the supremum over the ball near each $w_r$. Basically, since $w(t)$ depends on $W(0)$ and $A(0)$,
  directly working on $K(t)-K(0)$ is difficult. 
  The uniform bound \eqref{eqn:suptwK} allows us to
  overcome this dependence when applied to $K(t)-K(0)$.
\end{Remark}

Note that in this lemma we use $K(w)$ to indicate that the kernel $K$ is being evaluated at the weight vectors $w_r$ and ignore the time index $t$. In this Lemma, we use $\Xt(w_r)$ to denote $\Xt_r$ evaluated at $w_r$.

\proof For simplicity of notation, we use $\sup_{\widetilde{w}}$ to represent the supremum in \eqref{eqn:suptwK}, and $\sup_{\widetilde{w}_r}$ to represent
$\sup_{\{\widetilde{w}_r:\|\tilde{w}_r -w_r\|\le R\}}$. To prove this lemma, we work on the Frobenius norm instead of the spectral norm. Let us first write
\[
  z_{ijr} = \1[\widetilde{w}_r^\top x_i \geq 0, \widetilde{w}_r^\top x_j \geq 0] -\1[\wT_r x_i \geq 0, \wT_r x_j \geq 0].
\]
Next,
\begin{align*}
\norm{K(\widetilde{w}) - K(w)}^2 \leq \norm{K(\widetilde{w}) - K(w)}_F^2
                                 &= \frac{1}{m^2} \sum_{i,j=1}^n \norm{\xT_ix_j \sum_{r=1}^m z_{ijr} a_r\aT_r}_F^2 \\
  &\leq \frac{1}{m^2} \left(\sum_{i, j=1}^n \norm[\Big]{\sum_{r=1}^m z_{ijr} a_r\aT_r}_F \right)^2.
\end{align*}
The last step follows from the fact that $|\xT_ix_j| \leq 1$ due to Cauchy-Schwartz. Therefore,
\begin{align*}
  \sup_{\wt} \norm{K(\widetilde{w}) - K(w)} &\leq \frac{1}{m} \sup_{\wt} \sum_{i, j=1}^n \norm[\Big]{\sum_{r=1}^m z_{ijr} a_r\aT_r}_F \\
                                              &\leq \frac{1}{m} \sup_{\wt} \sum_{i, j=1}^n \sum_{r=1}^m |z_{ijr}| \norm{a_r\aT_r}_F \\
                                              &\leq \frac{d}{m}  \sum_{i, j=1}^n \sum_{r=1}^m \sup_{\wt_r}|z_{ijr}|,
\end{align*}
since  $\norm{a_r\aT_r}_F =  \norm{a_r}^2 = d$. Now we take expectation over the random vector $w_r$'s on both sides:
\begin{align*}
  \E_{w}[\sup_{\wt_r} \norm{K(\widetilde{w}) - K(w)}] &\leq \frac{d}{m}  \sum_{i, j=1}^n \sum_{r=1}^m \E_{w}[\sup_{\wt_r}  |z_{ijr}|].
\end{align*}
Next, we bound $\E_{w}[\sup_{\wt_r}  |z_{ijr}|]$. By definition of $z_{ijr}$,
\begin{align*}
  \label{eqnBoundzijr}
  |z_{ijr}| &= |\1[\wT_r x_i \geq 0, \wT_r x_j \geq 0] - \1[\widetilde{w}_r^\top x_i \geq 0, \widetilde{w}_r^\top x_j \geq 0]| \\
            &\leq |\1[\wT_r x_i \geq 0] - \1[\widetilde{w}_r^\top x_i \geq 0]| + |\1[\wT_r x_j \geq 0] - \1[\widetilde{w}_r^\top x_j \geq 0]| \\
            &\leq \1[|\wT_r x_i| \leq R] + \1[|\wT_r x_j| \leq R]. \numberthis
\end{align*}
The last step follows from the results in \citep[Lemma 3.2]{du2018_gradient}. So we get
\begin{align*}
  \E_{w}[\sup_{\wt_r}  |z_{ijr}|] &\leq \E_{w}[\1[|\wT_r x_i| \leq R] + \1[|\wT_r x_j| \leq R]] \\
  &= 2\Prob_{z \sim \normal(0, 1)}[|z| < R] \leq \frac{4R}{\sqrt{2\pi}} < 2R.
\end{align*}
Therefore,
\begin{equation*}
  \E_{w}[\sup_{\wt} \norm{K(\widetilde{w}) - K(w)}] < 2n^2dR.
\end{equation*}
Finally, by Markov's inequality, with probability at least $1-\delta$:
\begin{equation*}
  \sup_{\wt} \norm{K(\widetilde{w}) - K(w)} < \frac{2n^2dR}{\delta}.
\end{equation*}
\qedhere

\begin{Corollary}
  \label{corMinEigenKt_Case1}
  Suppose $\norm{w_r(t) - w_r(0)} \leq R \triangleq \frac{\lambda_0\delta}{8n^2d }$ for all $r \in [m] $ and $t \ge 0$ with probability at least $1-\delta$. We have
  \[
  \lambda_{\min}(K(t))  > \frac{\lambda_0}{2}
\]
with probability at least $1-3\delta$
if $m \ge C\frac{\lambda_n^2d\log(nd/\delta)}{\lambda^2_0}$.
\end{Corollary}
\proof
This is the direct consequence of Lemma \ref{lmMinEigenK0_Case1} and Lemma \ref{lmBoundKDiff_Case1}.
Since $\norm{w_r(t) - w_r(0)} \leq R = \frac{\lambda_0\delta}{8n^2d }$ with probability at least $1-\delta$ for all $t \geq 0$, then
\[
  \norm{ K(t) - K(0)}
  < 2n^2dR\delta = \frac{\lambda_0}{4}
\]
with probability at least $1-2\delta$. Using Weyl's inequality, we can bound:
\[
  \lambda_{\min}(K(t)) \geq \lambda_{\min}(K(0)) - \norm{K(t) - K(0)} > \lambda_0/2
\]
with probability at least $1-3\delta$
if $m \ge C\frac{\lambda_n^2d\log(nd/\delta)}{\lambda^2_0}$
as stated in Lemma \ref{lmMinEigenK0_Case1}.

\qedhere

In what follows, we show that $\norm{w_r(t) - w_r(0)} \leq R$ with high probability if $m$ is sufficiently large.
\begin{Lemma}
  Fix $t > 0$.  Suppose $\lambda_{\min}(K(s)) \geq \lambda_0/2$ for all $0 \le s < t$. Then, 
  \[
  \norm{X - U(s)}_F^2 \le \exp\left(-\frac{\lambda_0 s}{d}\right) \norm{X - U(0)}_F^2.
\]
Also, for each $r = 1, 2, \dots, m$:
\[
  \norm{w_r(t) - w_r(0)} \le \frac{d\sqrt{\lambda_n}\norm{X - U(0)}_F}{\sqrt{m}\lambda_0} \triangleq R'.
\]
\label{lmMovementInTime_Case1}
\end{Lemma}
\proof
For all $s \in [0, t)$, we have
\begin{align*}
  \frac{d}{ds} \norm{\vec(X - U(s))}_2^2  &= -2\vec(X - U(s))^\top \frac{1}{d} K(s)\vec(X - U(s)) \nonumber \\
                &\le -\frac{2}{d}\lambda_{\min}(K(s))\norm{\vec(X - U(s))}^2 \\
                &\le -\frac{\lambda_0}{d}\norm{X - U(s)}_F^2
\end{align*}
by the assumption $\lambda_{\min}(K(s)) \ge \lambda_0/2$. Therefore, the loss at time $s$ is upper-bounded by
\begin{align*}
  \label{eqnLossDecayWeakly}
  \norm{X - U(s)}_F^2 &= \norm{\vec(X - U(s))}^2 \\ &\le \exp \Bigl(-\frac{\lambda_0s}{d} \Bigr) \norm{\vec(X - U(0))}^2 \\
                                               &\le \exp \Bigl (-\frac{\lambda_0s}{d} \Bigr) \norm{X - U(0)}_F^2, \numberthis
\end{align*}
which decays exponentially with time $s$ at rate $\lambda_0/d$. 

To upper bound the movement of the weights $\norm{w_r(t) - w_r(0)}$, we use the above result while expanding the derivative of $w_r(s)$ over time $0 \le s < t$:
\begin{align*}
  \norm[\bigg]{\frac{\der}{\der s}w_r(s)} &= \norm[\bigg]{-\nabla_{w_r}L(W(s))} \\
                                   &= \norm[\bigg]{\frac{1}{\sqrt{md}}\sum_{i=1}^n\1[\wT_rx_i \ge 0]x_i\aT_r(x_i - u_i(s))} \\
  &= \norm[\bigg]{\frac{1}{\sqrt{md}}\Xt_r (X - U(s))^\top a_r} \\
      &\le \frac{\norm{X}\norm{a_r}}{\sqrt{md}} \norm{X - U(s)}_F \\
                                   &\leq \sqrt{\frac{\lambda_n}{m}}  \exp\Bigl(-\lambda_0s/d \Bigr) \norm{X - U(0)}_F,
\end{align*}
where the last step follows from $\norm{a_r}^2 = d, \norm{X}^2 = \lambda_n$ and Eq. \eqref{eqnLossDecayWeakly}. 
From the differential equation, $w_r(s)$ is continuous for all $s \in [0, t)$, and so is $\norm{w_r(s) - w_r(0)}$. Consequently, we can take the limit for $t' \rightarrow t$:
\begin{align*}
\norm{w_r(t) - w_r(0)}_2 &= \lim_{t' \rightarrow t} \norm{w_r(t') - w_r(0)}_2 \le \lim_{t' \rightarrow t} \int_{0}^{t'}  \norm[\bigg]{\frac{\der}{\der s}w_r(s)}ds \\ &\le \lim_{t' \rightarrow t} \int_{0}^{t'} \frac{\sqrt{\lambda_n}\exp\Bigl(-\lambda_0s/d \Bigr)\norm{X - U(0)}_F}{\sqrt{m}}ds \\
&\le \frac{d\sqrt{\lambda_n}\norm{X - U(0)}_F}{\sqrt{m}\lambda_0} \triangleq R',
\end{align*}
since $\exp\bigl(-\lambda_0s/d \bigr)$ is continuous at $s=t$. Therefore, we finish the proof. 

\qedhere

\begin{Lemma}
If $R' < R$, then $\lambda_{\min}(K(t)) \geq \frac{1}{2}\lambda_0$
for all $t \geq 0$. Moreover, 
  $
    \norm{w_r(t) - w_r(0)} \leq R'
  $
  and
  $
    \norm{X - U(t)}_F^2 \le \exp(-\frac{\lambda_0 t}{d}) \norm{X - U(0)}_F^2
  $
for all $r \in [m]$.
  \label{lmConvergence_Case1}
\end{Lemma}
\proof

We will prove this by contradiction. Assume the conclusion does not hold, meaning there exists $t_0$ such that:
\begin{equation*}
  t_0 = \inf \left \{ t > 0: \lambda_{\min}(H (t) ) \leq \lambda_0/2 \right \}.
\end{equation*}
We will argue that $t_0 > 0$ using the continuity. Since $w_r(t)$ is continuous in $t$, $K(t)$ and $\lambda_{\min}(K(t))$ are also continuous. Therefore, there exists $t' > 0$ such that for any $0 < \epsilon < \lambda_0/4$ we have
\[
  \lambda_{\min}(K(t')) > \lambda_{\min}(K(0)) - \epsilon > \lambda_0/2.
\]

Since $t_0 > 0$, then for any $0 \le s < t_0$, $\lambda_{\min}(H (s) ) \ge \lambda_0/2$. By Lemma \ref{lmMovementInTime_Case1}, we have for all $r \in [m]$:
\[
  \norm{w_r(t_0) - w_r(0)} \leq R' < R.
\]
Corollary \ref{corMinEigenKt_Case1} implies that $\lambda_0(H(t_0)) > \lambda_0/2$, which is a contradiction.

Therefore, we have proved the first part. For the second part, we have for all $t \geq 0$, $\lambda_{\min}(K(t)) \geq \frac{1}{2}\lambda_0$ and it follows from Lemma \ref{lmMovementInTime_Case1} that:
$
    \norm{w_r(t) - w_r(0)} \leq R'
$
for all $r \in [m]$ and
$
\norm{X - U(t)}_F^2 \le \exp(-\frac{\lambda_0 t}{d}) \norm{X - U(0)}_F^2.
$

\qedhere

Now, we bound $\norm{X - U(0)}_F$ to upper bound $R'$.
\begin{Claim}
  For any $\delta \in (0, 1)$, then $\norm{X - U(0)}_F^2 \leq \frac{2n}{\delta}$ with probability at least $1-\delta$.
  \label{clInitialLoss_Case1}
\end{Claim}
\proof We prove this using Markov's inequality. We use the independence between $A(0)$ and $W(0)$ to derive expressions for the expectation. In this proof, the expectations are evaluated over $W(0)$ and $A(0)$.
\begin{align*}
  \E[ \norm{X - U(0)}_F^2] &= \norm{X}_F^2 + \frac{1}{md}\E[\norm{A(0)\phi(W(0)^TX}_F^2] \\
                           &= n + \frac{1}{md}\E[\textrm{trace}(\phi(X^TW(0))A(0)\phi(W(0)^TX] \\
                           &= n + \frac{1}{md}\textrm{trace}(\E[\phi(X^TW(0))A(0)^\top A(0) \phi(W(0)^TX]) \\
                           &= n + \frac{1}{m}\textrm{trace}(\E[\phi(X^TW(0))\phi(W(0)^TX]) \\
                           &= n + \sum_{i=1}^n\E_w[\phi(\wT x_i)^2] \\
                           &= n + n\E_{z \in \normal(0, 1)}[z^2\1[z \geq 0]]= \frac{3n}{2},
\end{align*}
where in the fourth step we use $\E[A(0)^\top A(0)] = dI$, and in the last step we use the independence of the columns of $W(0)$. Using Markov, we get:
\[
  \norm{X - U(0)}_F^2 \leq \frac{2n}{\delta}
\]
with probability at least $1-\delta$.
\qedhere

\proof[Proof of Theorem~\ref{thmGradFlow_Case1}] If the following condition holds
\[
R' = \frac{d\sqrt{\lambda_n}\norm{X - U(0)}_F}{\sqrt{m}\lambda_0} \leq R = \frac{\delta\lambda_0}{8n^2d},
\]
then Lemma \ref{lmMovementInTime_Case1} follows. Using the condition with the bound $\norm{X - U(0)}_F \leq \sqrt{2n/\delta}$ in Claim \ref{clInitialLoss_Case1}, we obtain $m = \Omega\left( \frac{n^5d^4\lambda_n}{\lambda_0^4\delta^3}\right)$. This bound dominates the order of $m$ required for the concentration of $K(0)$ in the Corollary \ref{corMinEigenKt_Case1}, and therefore Theorem~\ref{thmGradFlow_Case1} follows.
\qedhere

\subsection{Gradient descent }
\label{sec:gradient-descent}

The above result for gradient flow can be viewed as a convergence rate for gradient descent in the weakly-trained regime with infinitesimally small step size. We now derive a convergence rate for gradient descent with finite step sizes. 

\begin{Theorem}
  \label{thmGradDescent_Case1}
  Suppose Assumptions~\ref{astUnitNorm} and \ref{astMinEigen} hold. The initial weights are independently drawn such that $w_r \sim \normal(0, I)$ and $a_r \sim \mathrm{Unif}(\{\pm 1\}^d)$ for all $r \in [m]$. If $m \ge C\frac{n^5d^4\lambda_n}{\lambda_0^4\delta^3}$  for some large enough constant $C$, then with probability at least $1-\delta$ the gradient descent on $W$ with step size $\eta = \Theta(\frac{\lambda_0}{nd\lambda_n})$,
  \begin{equation}
    \label{eqnGDConvergence_Case1}
        \norm{X - U(k)}_F^2 \leq \left(1-\frac{\eta \lambda_0}{2d}\right)^k \norm{X - U(0)}_F^2
      \end{equation}
      for $k=0, 1, \dots$
\end{Theorem}

We will prove Theorem \ref{thmGradDescent_Case1} by induction. The base case when $k = 0$ is trivially true. Assume Eq. \eqref{eqnGDConvergence_Case1} holds for $k'=0, 1, \dots, k$, then we show it holds for $k' = k+1$. To this end, we first prove $\norm{w_r(k+1) - w_r(0)}$ is small enough; then we use that property to bound $\norm{X - U(k+1)}_F^2$.

\begin{Lemma}
  \label{lmMovementInstep_Case1}
  If \eqref{eqnGDConvergence_Case1} holds for $k'=0, 1, \dots k$, then we have for all $r \in [m]$,
  \begin{align*}
    \norm{w_r(k+1) - w_r(0)} \leq \frac{4d\sqrt{\lambda_n}\norm{X - U(0)}_F}{\sqrt{m} \lambda_0} \triangleq R'.
  \end{align*}
\end{Lemma}
\proof We use the expression of the gradient in \eqref{eqnGradLossOverWr_Case1}, which is:
\begin{align*}
  \nabla_{w_r}L(W(k)) & = -\sum_{i=1}^n \frac{1}{\sqrt{md}}\1[w_r(k)^Tx_i \ge 0]x_i\aT_r (x_{i} - u_{i}(k)) \\ & = - \frac{1}{\sqrt{md}}\Xt_r(k)(X - U(k))^\top a_r.
\end{align*}
Then, the difference of the weight vector $w_r$ is:
\begin{align*}
  \norm{w_r(k+1) - w_r(0)}  &= \eta \norm[\Big]{\sum_{k'=0}^k\nabla_{w_r}L(w_r(k'))} \\
                           &= \eta \norm[\bigg]{\sum_{k'=0}^k\frac{1}{\sqrt{md}}\Xt_r(k')(X - U(k'))^\top a_r} \\
                             &\le \eta \frac{\norm{X}}{\sqrt{md}}\sum_{k'=0}^k\norm{\vec(X - U(k'))}_F\norm{a_r} \\
                           &\le \eta \frac{\sqrt{\lambda_n}}{\sqrt{m}}\sum_{k'=0}^k\Bigl(1 - \frac{\eta \lambda_0}{2} \Bigr)^{k'/2}\norm{X - U(0)}_F \\
                           &\le \eta \frac{\sqrt{\lambda_n}}{\sqrt{m}}\norm{X - U(0)}_F\sum_{k'=0}^\infty\Bigl(1 - \frac{\eta \lambda_0}{2d} \Bigr)^{k'/2} \\
                           &= \eta \frac{\sqrt{\lambda_n}}{\sqrt{m}}\norm{X - U(0)}_F \frac{1}{\eta \lambda_0/(4d)} \\
                           &= \frac{4d\sqrt{\lambda_n}\norm{X - U(0)}_F}{\sqrt{m}\lambda_0},
\end{align*}
where the third step and the fourth step follow from the facts that $\norm{\Xt_r(k')} \leq \norm{X} = \sqrt{\lambda_n}$ and $\norm{a_r} = \sqrt{d}$. The last step follows because $\sum_{i=0}^\infty (1-\eta\lambda_0/2)^{i/2}  \le \frac{4d}{\eta \lambda_0}$.

\qedhere

Now, let us derive the form of $X - U(k+1)$. First, we compute the difference of the prediction between two consecutive steps, similar to deriving $\frac{du_i(t)}{dt}$. For each $i \in [n]$, we have
\begin{align*}
  \label{eqnDiffui_Case1}
  &u_i(k+1) - u_i(k)  = \frac{1}{\sqrt{md}} \sum_{r=1}^m a_r \left( \phi( w_r(k+1)^\top x_i ) - \phi(w_r(k)^\top x_i ) \right) \\
                    &= \frac{1}{\sqrt{md}} \sum_{r=1}^m a_r \left( \phi \left( \Big( w_r(k) - \eta \nabla_{w_r}L(W(k)) \Big)^\top x_i \right) - \phi ( w_r(k)^\top x_i ) \right). \numberthis
\end{align*}
We split the right hand side into two parts: $v_{1,i}$ represents the terms that the activation pattern does not change and $v_{2,i}$ represents the remaining term that pattern may change. Formally speaking, for each $i \in [n]$, we define
\[
  S_i = \{r \in [m] : \1[ w_r(k+1)^\top x_i \geq 0] = \1[ w_r(k)^\top x_i \geq 0 ] \}, ~ \text{and } \Sc_i = [m] \backslash S_i.
\]

Then, we can formally define  $v_{1,i}$ and $v_{2,i}$ as follows:
\begin{align*}
&v_{1,i} \triangleq \frac{1}{ \sqrt{md} } \sum_{r \in S_i} a_r \left( \phi \left( \Big( w_r(k) - \eta \nabla_{w_r}L(W(k)) \Big)^\top x_i \right) - \phi( w_r(k)^\top x_i ) \right), \\
&v_{2,i} \triangleq \frac{1}{ \sqrt{md} } \sum_{r \in \Sc_i} a_r \left( \phi \left( \Big( w_r(k) - \eta \nabla_{w_r}L(W(k)) \Big)^\top x_i \right) - \phi( w_r(k)^\top x_i ) \right) .
\end{align*}
We write $v_1 = (v_{1,1}^\top, v_{1,2}^\top, \dots, v_{1,n}^\top)^\top$ and do the same for $v_2$, so
\begin{align*}
\vec(U(k+1) - U(k)) = v_1 + v_2 .
\end{align*}

In order to analyze $v_1 \in \R^n$, we define $K$ and $K^{\bot} \in \R^{nd \times nd}$  as follows:
\begin{align*}
K(k)_{i,j} = & ~ \frac{1}{m} \sum_{r=1}^m x_i^\top x_j \1[{ w_r(k)^\top x_i \geq 0, w_r(k)^\top x_j \geq 0 }] a_r\aT_r, \\
K(k)^{\bot}_{i,j} = & ~ \frac{1}{m} \sum_{r\in \Sc_i} x_i^\top x_j \1[ w_r(k)^\top x_i \geq 0, w_r(k)^\top x_j \geq 0]a_r\aT_r.
\end{align*}
Next, we write $\phi(z)= z\1[z \geq 0]$ to make use the definition of $S_i$ and expand the form of  $\nabla_{w_r}L(W(k))$:
\begin{align*}
  v_{1,i}
  &= \frac{1}{\sqrt{md}} \sum_{r \in S_i} a_r \Big(- \eta \nabla_{w_r}L(W(k)) \Big)^\top x_i \1[ w_r(k)^\top x_i \geq 0] \\
  &= \frac{\eta}{md} \sum_{j=1}^n \xT_ix_j \sum_{r \in S_i} \1[{ w_r(k)^\top x_i \geq 0 , w_r(k)^\top x_j \geq 0 }]a_r \aT_r  (x_j - u_j) \\
  &= \frac{\eta}{d} \sum_{j=1}^n  ( K_{i,j}(k) - K_{i,j}^{\bot}(k) ) (x_j - u_j),
\end{align*}
Then, we can write $v_1$ as:
\begin{align}
  \label{eqnV1compact}
  v_1 = \frac{\eta}{d} (K(k) - K^{\bot}(k)) \vec(X - U(k)),
\end{align}
and expand $\| X - U(k+1) \|_F^2$:
\begin{align*}
  \norm{ X - U(k+1)}_F^2
  &= \norm{ \vec(X - U(k+1))}^2 \\
  &= \norm{ \vec(X - U(k)) - \vec(U(k+1) - U(k)) }_F^2 \\
  &= \norm{X - U(k)}_F^2 - 2\vec(X - U(k))^\top \vec(U(k+1) - U(k)) \\
  &~~+ \norm{U(k+1) - U(k)}_F^2 .
\end{align*}
We can further expand the second term above using \eqref{eqnV1compact} as below:
\begin{align*}
  \vec(X - U(k))^\top &\vec(U(k+1) - U(k)) \\
  =&~ \vec( X - U(k) )^\top ( v_1 + v_2 )  \\
  =&~ \vec( X - U(k) )^\top v_1 + \vec( X - U(k) )^\top v_2   \\
  =&~ \frac{\eta}{d} \vec( X - U(k) )^\top K(k) \vec( X - U (k) ) - \frac{\eta}{d} \vec( X - U(k) )^\top K(k)^{\bot} ( X - U(k) ) \\ &~+ \vec( X - U(k) )^\top v_2.
\end{align*}
We define and bound the following quantities and bound them in Claims~\ref{clmC1_Case1}, \ref{clmC2_Case1}, \ref{clmC3_Case1} and \ref{clmC4_Case1}.
\begin{align*}
C_1 = & ~ -\frac{2\eta}{d} \vec( X - U(k) )^\top K(k) \vec( X - U (k) ) , \\
C_2 = & ~ \frac{2\eta}{d} \vec( X - U(k) )^\top K(k)^{\bot} ( X - U(k) ) , \\
C_3 = & ~ -2\vec( X - U(k) )^\top v_2 , \\
C_4 = & ~ \norm{U(k+1) - U(k)}_F^2 . 
\end{align*}

\proof[Proof of Theorem \ref{thmGradDescent_Case1}] We are now ready to prove the induction hypothesis. What we need to is to prove
\[
  \norm{X - U(k')}_F^2 \leq (1-\frac{\eta \lambda_0}{2d})^{k'} \norm{X - U(0)}_F^2
\]
holds for $k'=k+1$ with probability at least $1-\delta$. In fact,
\begin{align*}
\norm{ X - U(k+1) }_F^2 
&= \norm{ X - U(k) }_F^2 + C_1 + C_2 + C_3 + C_4 \\
&\leq  \norm{ X - U(k) }_F^2 \left( 1 - \frac{\eta \lambda_0}{d} + 8 \eta n R  + 8 \eta n R  + \eta^2 n\lambda_n \right) , 
\end{align*}
with probability at least $1-\delta$ where the last step follows from Claim~\ref{clmC1_Case1}, \ref{clmC2_Case1}, \ref{clmC3_Case1}, and \ref{clmC4_Case1}.

\paragraph{Choice of $\eta$ and $R$.}

We need to choose $\eta$ and $R$ such that
\begin{align}\label{eq:choice_of_eta_R}
( 1 - \frac{\eta \lambda_0}{d} + 8 \eta n R  + 8 \eta n R  + \eta^2 n\lambda_n ) \leq 1-\frac{\eta\lambda_0}{2d} .
\end{align}

If we set $\eta=\frac{\lambda_0 }{4nd\lambda_n}$ and $R=\frac{\lambda_0}{64nd}$, we have 
\begin{align*}
8 \eta n R  + 8 \eta n R =16\eta nR \leq  \frac{\eta \lambda_0} {4d} ,
\mathrm{~~~and~~~} \eta^2 n\lambda_n \leq  \frac{\eta \lambda_0}{4d}.
\end{align*}
Finally, 
\begin{align*}
\norm{ X - U(k+1) }_F^2 \leq \left( 1 - \frac{\eta \lambda_0}{2d} \right) \norm{ X - U(k) }_F^2 \cdot 
\end{align*}
holds with probability at least $1-\delta$ if $2n\exp(-mR) \leq \delta/3$.

\paragraph{Lower bound on the level of over-parameterization $m$.}

We require for any $\delta \in (0, 1)$ that
\begin{align*}
R'= \frac{4d\sqrt{\lambda_n}\norm{ X - U(0) }_F}{\sqrt{m}\lambda_0} < R = \min\left\{ \frac{\lambda_0}{64nd}, \frac{\lambda_0 \delta}{2n^2d} \right\},
\end{align*}
where the first bound on $R$ comes from the gradient descent whereas the second is required in Lemma \ref{lmBoundKDiff_Case1}. By Claim \ref{clInitialLoss_Case1} that $\norm{X - U(0)}_F \leq \sqrt{\frac{2n}{\delta}}$ with probability at least $1-\delta$, then we require
\[
  m \geq C\frac{n^5\lambda_nd^4}{\lambda^4_0\delta^3} ,
\]
for a sufficiently large constant $C>0$ so that the descent holds with probability $1-\delta$.

\qedhere

We give proofs for Claims \ref{clmC1_Case1}, \ref{clmC2_Case1}, \ref{clmC3_Case1}, and \ref{clmC4_Case1} in Appendix \ref{appdx:weakly_train}.


\section{Jointly-trained Autoencoders}
In the previous section, we analyzed the gradient dynamics of a two-layer autoencoder under the {weakly-trained} regime. We now analyze the jointly-trained regime where the loss is optimized over both sets of layer weights. For consistency of our presentation, we reuse some key notations in this section; for example, $K(t), U(t)$ have the same interpretation as before but possess a different closed form.

\subsection{Gradient flow}
\label{sec:gradflow_joint}
The loss function we consider for this \emph{jointly-trained} regime is the same:
\begin{equation}
\label{eqnEmpLoss_joint}
   L(W, A) = \frac{1}{2}\sum_{i=1}^n\norm{x_i - \frac{1}{\sqrt{md}}A\phi(W^\top x_i)}^2.
\end{equation}
The difference is that the optimization is now taken over \emph{both} weights $W$ and $A$. To make the comparison easier, the matrices $W$ and $A$ are randomly initialized in the same way such that
\[
  w_{ij}(0) \sim \mathcal{N}(0, 1), ~ a_{ij}(0) \sim \mathrm{Unif}(\{- 1, 1\})
\]
are drawn independently for each pair $(i, j)$. $W$ and $A$ are then updated using gradient descent with  step size $\eta$:
\begin{align}
  W(k+1) &= W(k) - \eta \nabla_W L(W(k), A(k)), ~k=0, 1, \dots \label{eqnUpdateWJoint} \\
  A(k+1) &= A(k) - \eta \nabla_A L(W(k), A(k)), ~k=0, 1, \dots \label{eqnUpdateAJoint}
\end{align}
Similar to the previous case, we derive the gradients of $L(W, A)$ with respect the column $w_r$ of $W$ and $a_r$ of $A$. The gradient $\nabla_{w_r}L(W, A)$ is the same in \eqref{eqnGradLossOverWr_Case1} in Section \ref{sec:gradflow_case1} whereas $\nabla_{a_r}L(W, A)$ is standard:
\begin{align}
  \nabla_{w_r}L(W, A)
                &= - \frac{1}{\sqrt{md}} \sum_{i=1}^n \1[w_r^Tx_i \ge 0]x_i\aT_r  (x_{i} - u_{i}), \numberthis
                  \label{eqnGradLossOverWr_joint} \\
  \nabla_{a_r}L(W, A) &= -\frac{1}{\sqrt{md}}\sum_{i=1}^n\phi(\wT_rx_i) (x_{i} - u_{i}). \numberthis
                  \label{eqnGradLossOverAr_joint}
\end{align}
Consider two ODEs, one for each weight vector over the continuous time $t$:
\begin{align}
  \frac{\df w_r(t)}{\dt} &= -\nabla_{w_r}L( W(t), A(t) ), \label{eqnGradODEWr_joint} \\
  \frac{\df a_r(t)}{\dt} &= -\nabla_{a_r}L( W(t), A(t) ). \label{eqnGradODEAr_joint}
\end{align}

Using~\eqref{eqnGradLossOverWr_joint}, \eqref{eqnGradLossOverAr_joint},  \eqref{eqnGradODEWr_joint} and \eqref{eqnGradODEAr_joint}, the continuous-time dynamics of the predicted output, $u_i(t)$, for sample $x_i$ is given by:
\begin{align*}
  \frac{\df u_{i}(t)}{\dt} &= \frac{\df}{\dt} \left( \frac{1}{\sqrt{md}}\sum_{r=1}^m a_r\phi(\wT_rx_{i}) \right) \\
                    &= \frac{1}{\sqrt{md}} \sum_{r=1}^m \left( J_{w_r} (a_r\phi(\wT_rx_{i}))\frac{dw_r}{dt} + J_{a_r} (a_r\phi(\wT_rx_{i}))\frac{da_r}{dt} \right) \\
                    &= \frac{1}{\sqrt{md}} \sum_{r=1}^m \left( 1[w_r^Tx_i \ge 0]\aT_rx_i(-\nabla_{w_r}L(W, A) ) + \phi(\wT_rx_i) (-\nabla_{a_r}L(W, A) )  \right) \\
                      &= - \frac{1}{\sqrt{md}} \sum_{j=1}^n \sum_{r=1}^m \left( 1[w_r^Tx_i \ge 0, w_r^Tx_j \ge 0]\xT_ix_j a_r\aT_r + \phi(\wT_rx_i)\phi(\wT_rx_j) I \right) (x_{j} - u_{j}).
\end{align*}
In these expresssions, we skip the dependence of the weight vectors on time $t$ and simply write them as $w_r$ and $a_r$. Vectorizing $\frac{dU(t)}{dt}$, we get to the key equation that characterizes the dynamics of $U(t)$:
\begin{align*}
\label{eqnPredDynamics_joint}
  \frac{\df \vec(U(t))}{\dt} 
  &= \frac{1}{d} \Bigl( G(t) + H(t) \Bigr) \vec(X - U(t). \numberthis
\end{align*}
In the above equation, $G(t)$ is a size-$nd\times nd$ matrix of the form:
\begin{align}
  G(t) = \frac{1}{m}\sum_{r=1}^m \Xt_r(t)^\top \Xt_r(t) \otimes a_r(t)a_r(t)^\top,
  \label{eqnGt_joint}
\end{align}
where
$
  \Xt_r(t) = \Bigl[\1[w_r(t)^\top x_{1} \ge 0]x_{1}, \dots, \1[w_r(t)^\top x_{n} \ge 0]x_{n}\Bigr],
$
while $H(t)$ is a size-$nd\times nd$ matrix:
\begin{align}
  H(t) = \frac{1}{m}\sum_{r=1}^m\phi(X^\top w_r(t))\phi(w_r(t)^\top X) \otimes I.
  \label{eqnHt_joint}
\end{align}
Let us emphasize again that $G(t)$ is precisely the kernel that governs the dynamics for the {weakly-trained} case. On the other hand, $H(t)$ is a Kronecker form of the Hessian of the loss function derived with respect to $A$, using the features produced at the output of the $\relu$ activations. 

As shown in Section \ref{sec:kernels}, assuming randomness and independence of $W(0)$ and $A(0)$, we can prove that as $m \rightarrow \infty$, $H(0)$ and $G(0)$ converge to the corresponding NTKs whose minimum eigenvalues are assumed to be positive. More specifically, we have
\begin{align*}
  G^\infty &= \E_{W(0), A(0)}[G(0)] \\ &= \E_{w(0), a(0)}[\Xt(0)^\top \Xt(0) \otimes a(0) a(0)^\top] \\ &= \E_{w}[\Xt^\top \Xt] \otimes I. \numberthis
                                      \label{eqnGinfty_joint}
\end{align*}
and 
\begin{align*}
  H^\infty &= \E_{W(0), A(0)}[H(0)] \\ &= \E_{w(0), a(0)}[\phi(X^\top w(0))\phi(w(0)^\top X)] \otimes I.
                                                                                                              \numberthis
  \label{eqnHinfty_joint}
\end{align*}
Denote the time-dependent kernel $ K(t) = G(t) + H(t) $. Since both $G(t)$ and $H(t)$ are positive semi-definite, we only focus on $H(t)$ for reasons that will become clear shortly. 

Since $G(t)$ is also positive definite with high probability (Section \ref{sec:gradflow_case1}), the flow convergence can be also boosted by the positive definiteness of $G^\infty$. By Assumption~\ref{astMinEigen},
\[
  \lambda_{\min}(K^\infty) \ge \lambda_{\min}(H^\infty) \ge \lambda_0 >0. 
\]
Since $G(0)$ is positive semi-definite, in order to bound the minimum eigenvalue of $K(0)$, all we need is to bound that of $H(0)$. Importantly, we observe that the smoothness of the kernel $H(t)$ is much better as a function of the deviation of the weights from the initialization. This allows the weights to change with a larger amount than merely using $G(t)$, and enables us to significantly reduce the number of parameters required for the gradient to reach a global optimum.

Our main result for gradient flow of the {jointly-trained} autoencoder is given by:

\begin{Theorem}[Linear convergence of gradient flow, jointly-trained regime]
  \label{thmGradFlow_joint}
  Suppose Assumptions~\ref{astUnitNorm} and \ref{astMinEigen} hold. The initial weights are independently drawn such that $w_r \sim \normal(0, I)$ and $a_r \sim \mathrm{Unif}(\{-1,1 \}^d)$ for all $r \in [m]$. If $m \ge C\frac{n\lambda_n^3d}{\lambda_0^4\delta^2}$  for some large enough constant $C$, then with probability at least $1-\delta$,
  \[
    \norm{X - U(t)}_F^2 \leq \exp\bigl(-\frac{\lambda_0 t}{d}\bigr)\norm{X - U(0)}_F^2.
  \]
\end{Theorem}

\begin{Remark}
  \normalfont We initialize the second-layer weights $A$ with independent Rademacher entries. This is for convenience of analysis because such $A$ has constant-norm columns. However, similar results should easily follow for initialization with more practical schemes (for example, i.i.d.~Gaussians).
\end{Remark}

We will first state and prove a few auxiliary results in Lemmas \ref{lmMinEigenH0_joint}, \ref{lmBoundHDiff_joint}, and \ref{lmMinEigennMovementAnyTime_joint} and then use them to prove Theorem~\ref{thmGradFlow_joint}.

\begin{Lemma}
  \label{lmMinEigenH0_joint}
  For any $\delta \in (0, \frac{\lambda_0}{12d\lambda_n^2})$, if $m \ge C\frac{nd \lambda_n\log^2(nd/\delta)}{\lambda_0^2}$ for some large enough constant $C$, then with probability at least $1 - 1/(2nd)^{2\log nd} - m\delta$,
  one obtains $\norm{H(0) - H^\infty} \leq {\lambda_0}/{4}$ and $\lambda_{\min}(H(0)) \geq 3\lambda_0/4$ .
\end{Lemma}
The proof of this Lemma is deferred to Appendix \ref{appdx:jointly_train}.

\begin{Lemma}
   \label{lmBoundHDiff_joint}
    Suppose $\norm{W(t) - W(0)}_F \leq R_w$.
   Then,
 \[
    \norm{H(t) - H(0)} \leq \frac{\lambda_n}{m} (2\norm{W(0)} + R_w)R_w.
  \]
  Particularly, 
  if $\frac{\lambda_n}{m} (2\norm{W(0)} + R_w)R_w \le \frac{\lambda_0}{4}$, then $\norm{H(t) - H(0)} \leq \frac{\lambda_0}{4}$. Therefore, $\lambda_{\min} (K(t) > \frac{\lambda_0}{2}$ if $\lambda_{\min}(H(0)) \ge \frac{3\lambda_0}{4}$.
\end{Lemma}
\begin{Remark}
  \normalfont Let us compare with Lemma \ref{lmBoundKDiff_Case1}. Note that compared with that bound, $O(n^2dR_w)$ on the kernel perturbation, here the spectral norm bound on $H(t) - H(0)$ is significantly better in two ways: 
 
 \begin{enumerate}[label=(\roman*)] 
\item the bound scales with $1/\sqrt{m}$, which later determines the over-parameterization and 
\item the movement is now characterized by the total $\norm{W(t) - W(0)}_F$. This is possible due to the smoothness of the $\relu$ activation, which is the reason why we focus on $H(t)$ instead of $G(t)$. 
\end{enumerate}
\end{Remark}
\proof We apply the triangle inequality and use the Lipschitz property of the rectified linear unit to bound the difference. Recall that
\[
  H(t) = \frac{1}{m} \sum_{r=1}^m\phi(X^\top w_r(t))\phi(w_r(t)^\top X) = \frac{1}{m} \phi(X^\top W(t))\phi(W(t)^\top X).
\]
Then, we can upper bound the perturbation as follows:
\begin{align*}
  \norm{H(t) - H(0)} &\leq \frac{1}{m}\norm[\big]{\phi(X^\top W(t))\phi(W(t)^\top X) - \phi(X^\top W(0))\phi({W(0)}^{\top} X)} \\
                     &\leq \frac{1}{m}\norm{\phi(X^\top W(t)}\norm{\phi(W(t)^\top X) - \phi(W(0)^\top X)} \\ &~~+ \frac{1}{m} \norm{\phi(X^\top W(t)) - \phi(X^\top W(0))}\norm{\phi({W(0)}^{\top} X)} \\
                     &\leq \frac{1}{m}\norm{X}^2\left(\norm{W(t)} + \norm{W(0)} \right)\norm{W(t) - W(0)}_F \\
                     &\leq \frac{\lambda_n}{m}\left(2\norm{W(0)} + \norm{W(t) - W(0)}\right)\norm{W(t) - W(0)}_F \\
                     &\leq \frac{\lambda_n}{m} (2\norm{W(0)} + R_w)\lambda_nR_w.
\end{align*}
In the third step, we use the fact that the $\relu$ function is 1-Lipschitz and $\norm{\phi(\XT W(t)} \leq \norm{X}\norm{W(t)}$. The last step follows by $\norm{W(t) - W(0)}_F \le R_w$.

Using the condition and Weyl's inequality, one can easily show that $\lambda_{\min}(K(t)) \geq \lambda_{\min}(H(t)) > \lambda_0/2$.

\qedhere

We haved proved that as long as the weight matrix $W(t)$ do not change much over $t$, the minimum eigenvalue of $K(t)$ stays positive. Next, we show that this implies the exponential decay of the loss with iteration, and give a condition under which the weights do not change much.

\begin{Lemma}
  Fix $t >0$.  Suppose $\lambda_{\min}(K(s)) \ge \frac{\lambda_0}{2}$
  for all $0 \le s < t$.
  Then 
  \[
  \norm{X - U(s)}_F^2 \le \exp\Bigl(-\frac{\lambda_0s}{d}\Bigr) \norm{X - U(0)}_F^2.
\]
\label{lmConvergenceFlow_joint}
\end{Lemma}
\proof
We have $\lambda_{\min}(K(s)) \geq \frac{\lambda_0}{2}$, then
\begin{align*}
                 \frac{\der}{\der s} \left( \norm{\vec(X - U(s))}_2^2 \right)
                      &= -2\vec(X - U(s))^\top \frac{K(t)}{d}\vec(X - U(s)) \nonumber \\
                &\le -\frac{2}{d}\lambda_{\min}(K(s))\norm{\vec(X - U(s))}_2^2 \\
                &\le -\frac{\lambda_0}{d}\norm{X - U(s)}_F^2, \numberthis
\end{align*}
since $\lambda_{\min}(H(s)) \ge \frac{\lambda_0}{2}$. Therefore,
\begin{align*}
  \norm{X - U(s)}_F^2 = \norm{\vec(X - U(s))}^2 &\le \exp\Bigl(-\frac{\lambda_0s}{d} \Bigr) \norm{\vec(X - U(0))}^2 \\
                                               &\le \exp\Bigl(-\frac{\lambda_0s}{d} \Bigr) \norm{X - U(0)}_F^2.
\end{align*}
\qedhere

\begin{Lemma}
  Fix $t >0$. Suppose $\lambda_{\min}(K(s)) \ge \frac{\lambda_0}{2}$ and $\norm{A(s) - A(0)}_F \leq R_a$ for all $0 \le s < t$.
  For all $r \in [m]$, we have
\[
  \norm{W(t) - W(0)}_F \le \frac{2\sqrt{d\lambda_n} (\norm{A(0)} + R_a)\norm{X - U(0)}_F }{\sqrt{m}\lambda_0} \triangleq R'_w.
\]
\label{lmMovementInTimeWr_joint}
\end{Lemma}
\proof
For $s \in [0, t)$, we have
\begin{align*}
  \frac{\df}{\der s}w_r(s) &= -\nabla_{w_r}L(W(s), A(s) ) \\
  &= \frac{1}{\sqrt{md}} \sum_{i=1}^n \1[w_r^Tx_i \ge 0]x_ia_r(s)^\top  (x_{i} - u_{i} (s) ) \\
                                   &= \frac{1}{\sqrt{md}}\Xt_r (X - U(s))^\top_r a_r(s).
\end{align*}
Then, one can bound the entire weight matrix as follows:

\begin{align*}
  \norm[\bigg]{\frac{\df}{\df s}W(s)}_F
  &\leq \frac{\norm{X}}{\sqrt{md}} \norm[\big]{ (X - U(s))^\top A(s)  }_F \\
    &\le \frac{\sqrt{\lambda_n}}{\sqrt{md}} \norm{X - U(s)}_F \norm{A(s)}\\
    &\le \frac{\sqrt{\lambda_n} (\norm{A(0)} + R_a)}{\sqrt{md}}  \exp\Bigl(-\frac{\lambda_0s}{2d} \Bigr)\norm{X - U(0)}_F.
\end{align*}
In the second step, we use the fact $\norm{CD}_F \leq \norm{C}_F\norm{D}$ for any matrices $C, D$, and $\norm{X}^2 = \lambda_n$. The last step follows from $\norm{A(s)} \leq \norm{A(0)}  + R_a$ and Lemma \ref{lmConvergenceFlow_joint}. Using the same continuity, we have
\begin{align*}
  \norm{W(t) - W(0)}_F &\le 
  \lim_{t' \rightarrow t} \int_{0}^{t'}  \norm[\bigg]{\frac{\df}{\der s}W(s)}_F \\
  &\le \lim_{t' \rightarrow t}\int_{0}^{t'} \frac{\sqrt{\lambda_n} (\norm{A(0)} + R_a) \exp\Bigl(-\frac{\lambda_0s}{2d} \Bigr)\norm{X - U(0)}_F}{\sqrt{md}}ds \\
                                                                          &\le \frac{2\sqrt{\lambda_n} (\norm{A(0)} + R_a)  \norm{X - U(0)}_F}{\sqrt{md}} \lim_{t' \rightarrow t}\int_{0}^{t'} \exp\Bigl(-\frac{\lambda_0s}{2d} \Bigr)ds \\
&\le \frac{2\sqrt{d\lambda_n} (\norm{A(0)} + R_a) \norm{X - U(0)}_F}{\sqrt{m}\lambda_0} \triangleq R_w'.
\end{align*}

\qedhere

\begin{Lemma}
  Fix $t >0$.  Suppose $\lambda_{\min}(K(s)) \ge \frac{\lambda_0}{2}$ and $\norm{W(s) - W(0) }_F \leq R_w$ for all $0 \le s < t$,
then for $r = 1, 2, \dots, m$
\[
  \norm{A(t) - A(0)}_F \le \frac{2\sqrt{d\lambda_n} (\norm{W(0)} + R_w) \norm{X - U(0)}_F}{\sqrt{m}\lambda_0} \triangleq R'_a.
\]
\label{lmMovementInTimeAr_joint}
\end{Lemma}
\proof
For $s \in [0, t)$, we use the gradient derived in \eqref{eqnGradLossOverAr_joint} and \eqref{eqnGradODEAr_joint} to obtain:
\begin{align*}
  \frac{\df}{\dt s}a_r(s) &= -\nabla_{a_r}L( W(s), A(s) ) \\
  &= \frac{1}{\sqrt{md}}\sum_{i=1}^n\phi(\wT_rx_i) (x_{i} - u_{i}(s)) \\
  &= \frac{1}{\sqrt{md}} (X - U(s)) \phi(\XT w_r(s))
\end{align*}
Then, one can write
\begin{align*}
  \norm[\bigg]{\frac{\df}{\dt s}A(s)}_F
  &= \norm[\bigg]{\frac{1}{\sqrt{md}} (X - U(s)) \phi(\XT W(s))}_F \\
    &\le \frac{\sqrt{\lambda_n}}{\sqrt{md}} \norm{X - U(s)}_F\norm{W(s)}\\
    &\le \frac{\sqrt{\lambda_n}(\norm{W(0)} + R_w) }{\sqrt{md}} \exp\Bigl(-\frac{\lambda_0s}{2d} \Bigr)  \norm{X - U(0)}_F,
\end{align*}
where we use $\norm{X} \le \sqrt{\lambda_n}$, $\norm{W(s)} \leq \norm{W(0)} + R_w$. The
last step follows from Lemma \ref{lmConvergenceFlow_joint}. Now, we integrate out $s$:
\begin{align*}
\norm{A(t) - A(0)}_F \le \int_{0}^t  \norm[\bigg]{\frac{\df}{\dt s}A(s)}_F &\le \int_{0}^t \frac{\sqrt{\lambda_n}(\norm{W(0)} + R_w)\exp\Bigl(-\frac{\lambda_0s}{2d} \Bigr)\norm{X - U(0)}_F}{\sqrt{m}}ds \\
                                                                          &\le \frac{2\sqrt{d\lambda_n}(\norm{W(0)} + R_w)\norm{X - U(0)}_F}{\sqrt{m} \lambda_0} = R'_a,
\end{align*}
which is what we need.

\qedhere

\begin{Lemma}
  If $R'_w < R_w$ and $R'_a < R_a$, then for all $t \ge  0$, we have
  \begin{itemize}
  \item[(i)] $\lambda_{\min}(K(t)) \ge \frac{\lambda_0}{2}$; and for all $r \in [m]$, $\norm{W(t) - W(0) }_F \leq R'_w$,  $\norm{A(t) - A(0) }_F \leq R'_a$;
  \item[(ii)] If (i) holds, then $\norm{X - U(t)}_F^2 \le \exp\Bigl(-\frac{\lambda_0}{d}\Bigr) \norm{X - U(0)}_F^2.$
  \end{itemize}
  \label{lmMinEigennMovementAnyTime_joint}
\end{Lemma}
\proof
Suppose on the contrary that
\begin{equation}
  \label{eqEventT}
  \mathcal{T} = \left\{ t \ge 0:  \lambda_{\min}(K(t)) \le \frac{\lambda_0}{2} \text{ {or} } \norm{W(t) - W(0) }_F > R'_w \text{ {or} } \norm{A(t) - A(0) }_F > R'_a \right\}.
\end{equation}
is not an empty set. Therefore, $t_0\triangleq\inf\mathcal{T}$ exists. Using the same contuinity argument as in Lemma \ref{lmMovementInTime_Case1}, one can verify that $t_0 > 0$. First, if $\lambda_{\min}(K(t_0)) \leq \frac{\lambda_0}{2}$, then by Lemma \ref{lmBoundHDiff_joint}, $\| W(t_0) - W(0) \|_F > R_w > R_w'$, which is a contradiction because it violates the minimality of $t_0$.

The other two cases are similar, so we will prove one of them. If it holds true that
\[
  \norm{W(t) - W(0) }_F > R'_w.
\]
The definitions of $t_0$ and $\mathcal{T}$ implies that for any $s \in [0, t_0)$,
$\lambda_{\min}(K(s)) \ge \frac{\lambda_0}{2}$ and $\norm{A(s) - A(0) }_F \leq R'_a$. Then, Lemma \ref{lmMovementInTimeWr_joint} leads to:
\[
  \norm{W(t_0) - W(0) }_F \le R'_w,
\]
which is a contradiction. Therefore, we have finish the proof.

\qedhere

\proof[Proof of Theorem \ref{thmGradFlow_joint}] With the results, we can can prove the Theorem. From Lemma \ref{lmMinEigennMovementAnyTime_joint}, if $R'_w \leq R_w$ and $R'_a \le R_a$, then
\[
  \norm{X - U(t)}_F^2 \le \exp\Bigl(-\frac{\lambda_0}{d}\Bigr) \norm{X - U(0)}_F^2.
\]
We only need the conditions  $R_w'=R'_a \leq R_w = R_a$ to satisfy for this to work. The conditions are
\begin{align*}
  &\frac{\lambda_n}{m} (2\norm{W(0)} + R_w)R_w \leq \frac{\lambda_0}{4}; \\
  \text{ and } &R_w < R_w' = \frac{2\sqrt{d\lambda_n} (\norm{A(0)} + R_a)\norm{X - U(0)}_F }{\sqrt{m}\lambda_0}.
\end{align*}
Note that $\norm{X - U(0)}_F^2 \leq 3n/2\delta$ with probability at least $1-\delta$. Also, using a standard bound on sub-Gaussian matrices, we have $\norm{W(0)} \leq 2\sqrt{m} + \sqrt{d}$ and $\norm{A(0)} \leq 2\sqrt{m} + \sqrt{d}$ with probability at least $1 - 2\exp(-m)$. Then if
we have the order of $m \ge \Omega\left( \frac{nd\lambda_n^3}{\lambda_0^4\delta^2}\right)$. Therefore, we finished the proof for the gradient flow Theorem.

\qedhere

\subsection{Gradient descent}
\label{sec:descent2}

As above, we will now appropriately discretize the gradient flow to obtain a convergence result for gradient descent with finite step size for the jointly-trained regime.

\begin{Theorem}
  \label{thmGradDescent_joint}
  Suppose Assumptions~\ref{astUnitNorm} and \ref{astMinEigen} hold. At initialization, suppose the weights are independently drawn from $w_r \sim \normal(0, I)$ and $a_r \sim \mathrm{Unif}(\{-1, 1\}^d)$ for all $r \in [m]$. If $m \ge C\frac{n\lambda_n^3d}{\lambda_0^4\delta^2}$ for some large enough constrant $C$, then with probability at least $1-\delta$ the gradient descent on $W$ with step size $\eta = \Theta(\frac{\lambda_0}{n\lambda_n})$,
  \begin{equation}
    \label{eqnGDConvergence_joint}
        \norm{X - U(k)}_F^2 \leq (1-\frac{\eta \lambda_0}{2d})^k \norm{X - U(0)}_F^2.
  \end{equation}
\end{Theorem}

We will prove \ref{thmGradDescent_joint} by induction. The base case when $k = 0$ is trivially true. Assume holds for $k'=0, 1, \dots, k$ and we want to show \eqref{eqnGDConvergence_joint} for $k' = k+1$. First, we prove that $\norm{W(k+1) - W(0)}_F$ and $\norm{A(k+1) - A(0)}_F$ are small enough, and we then use that to bound $\norm{X - U(k+1)}_F^2$.

In this section, we define and assume that
  \[
    R_w < \frac{4\sqrt{d\lambda_n} (\norm{A(0)} + R_a) \norm{X - U(0)}_F}{\sqrt{m}\lambda_0} \triangleq R'_w \quad\mbox{and}\quad
    R_a < \frac{4\sqrt{d\lambda_n} (\norm{W(0)} + R_w) \norm{X - U(0)}_F}{\sqrt{m}\lambda_0} \triangleq R'_a.
  \]
First, we show the following auxiliary lemma. 
\begin{Lemma}
  \label{lmMovementInstep_joint}
  If the condition \eqref{eqnGDConvergence_joint} holds for $k'=0, 1, \dots k$, then we have
  \begin{align*}
    \norm{W(k+1) - W(0)}_F \leq R'_w, \text{ and } \norm{A(k+1) - A(0)}_F \leq R'_a
  \end{align*}
  with probability at least $1-\delta$ for any $\delta \in (0, 1)$.
\end{Lemma}

\proof We prove this by induction. Clearly, both hold when $k'=0$. Assuming that both hold for $k' \leq k$. We will prove both hold for $k'=k+1$.

We use the expression of the gradients over $w_r$ and $a_r$ in \eqref{eqnGradLossOverWr_joint} and \eqref{eqnGradLossOverAr_joint}:
\begin{align*}
  \nabla_{w_r}L(W(k), A(k) )  &= -\sum_{i=1}^n \frac{1}{\sqrt{md}}\1[w_r(k)^Tx_i \ge 0]x_ia_r(k)^\top (x_{i} - u_{i}(k)) \\
  &= - \frac{1}{\sqrt{md}}\Xt_r(k)(X - U(k)) a_r(k), \\
  \nabla_{a_r}L(W(k), A(k) ) &= -\sum_{i=1}^n \frac{1}{\sqrt{md}}\phi(w_r(k)^\top x_i) (x_{i} - u_{i}(k)) \\
                              &= - \frac{1}{\sqrt{md}}(X - U(k)) \phi(\XT w_r(k)).
\end{align*}
Then, the difference of the weight matrix $W$ is:
\begin{align*}
  \norm{W(k+1) - W(0)}_F 
                           &\le \eta \frac{\norm{X}}{\sqrt{md}}\sum_{k'=0}^k\norm{\vec(X - U(k'))}_F(\norm{A(0)} + R_a) \\
                           &\le \eta \frac{\sqrt{\lambda_n} (\norm{A(0)} + R_a) }{\sqrt{m}}\sum_{k'=0}^k\Bigl(1 - \frac{\eta \lambda_0}{2d} \Bigr)^{k'/2}\norm{X - U(0)}_F \\
                           &\le \eta \frac{\sqrt{\lambda_n} (\norm{A(0)} + R_a)}{\sqrt{m}}\norm{X - U(0)}_F\sum_{k'=0}^\infty\Bigl(1 - \frac{\eta \lambda_0}{2d} \Bigr)^{k'/2} \\
                           &= \eta \frac{\sqrt{\lambda_n} (\norm{A(0)} + R_a)}{\sqrt{m}}\norm{X - U(0)}_F \frac{4d}{\eta \lambda_0} \\
                           &= \frac{4\sqrt{d\lambda_n} (\norm{A(0)} + R_a) \norm{X - U(0)}_F}{\sqrt{m}\lambda_0} = R_w',
\end{align*}
where the third step and fourth step follow from $\norm{\Xt_r(k')} \leq \norm{X} = \sqrt{\lambda_n}$ and the induction hypothesis $\norm{A(k')} \leq \norm{A(0)} + R_a$. The last step follows from
\begin{align*}
  \sum_{i=0}^\infty (1-\eta\lambda_0/2)^{i/2}
  \le \frac{4d}{\eta \lambda_0}.
\end{align*}
  
Similarly, we bound the difference of the weight matrix $A$ between time $k+1$ and $0$:
\begin{align*}
  \norm{A(k+1) - A(0)}_F 
                           &= \eta \norm[\bigg]{\sum_{k'=0}^k\frac{1}{\sqrt{md}}(X - U(k') ) \phi(\XT W(k') )}_F \\
                             &\le \eta \frac{\norm{X}}{\sqrt{md}}\sum_{k'=0}^k\norm{X - U(k') }_F\norm{W(k')} \\
                           &\le \eta \frac{\sqrt{\lambda_n}}{\sqrt{md}}\sum_{k'=0}^k\Bigl(1 - \frac{\eta \lambda_0}{2d} \Bigr)^{k'/2}\norm{X - U(0) }_F (\norm{W_0} + R_w) \\
                           &= \eta \frac{\sqrt{\lambda_n} (\norm{W_0} + R_w) \norm{X - U(0)}_F}{\sqrt{m}}  \frac{4}{\eta \lambda_0}\\
                           &=  \frac{4\sqrt{\lambda_n} (\norm{W_0} + R_w) \norm{X - U(0)}_F}{\sqrt{m} \lambda_0} = R'_a
\end{align*}
where the third step and fourth step follow from the facts that $\norm{\phi(\XT W(k'))} \leq \norm{X}\norm{W(k')}$ and $\norm{X} = \sqrt{\lambda_n}$, and $\norm{W(k')} \le \norm{W(0)} + R_w$.

We have therefore shown that $\norm{W(k') - w(0)}_F \leq R_w'$ and $\norm{A(k') - A(0)}_F \leq R_a'$ for $k' = k+1$. 
\qedhere

Now, we expand $\norm{ X - U(k+1)}_F^2$ in terms of the step $k$. Recall the update rule in \eqref{eqnUpdateWJoint} and \eqref{eqnUpdateAJoint} that
\begin{align}
  W(k+1) &= W(k) - \eta \nabla_W L(W(k), A(k)), ~k=0, 1, \dots \\
  A(k+1) &= A(k) - \eta \nabla_A L(W(k), A(k)), ~k=0, 1, \dots
\end{align}
where the gradients is given above. Next, we compute the difference of the prediction between two consecutive steps, a discrete version of $\frac{du_i(t)}{dt}$.  For each $i \in [n]$, we have 
\begin{align*}
  \label{eqnDiffui_Case1}
  u_i(k+1) &- u_i(k) = \frac{1}{\sqrt{md}} \sum_{r=1}^m \left( a_r(k+1) \phi( w_r(k+1)^\top x_i ) - a_r(k)\phi(w_r(k)^\top x_i ) \right) \\
                    &= \frac{1}{\sqrt{md}} \sum_{r=1}^m \left( \bigl ( a_r(k) - \eta \nabla_{a_r} L \bigr) \phi\left( \bigl(w_r(k) - \eta \nabla_{w_r} L \bigl)^\top x_i \right) - a_r(k)\phi(w_r(k)^\top x_i ) \right).
                        \numberthis
\end{align*}
For a particular $r$, if the activation pattern does not change, we can write the inside term as:
\begin{align*}
  &\bigl ( a_r(k) - \eta \nabla_{a_r} L \bigr) \phi\left( \bigl(w_r(k) - \eta \nabla_{w_r} L \bigl)^\top x_i \right) - a_r(k)\phi(w_r(k)^\top x_i ) \\
  &= \left( -\eta a_r(k) \left( \nabla_{w_r} L  \right)^\top - \eta ( \nabla_{a_r} L ) w_r(k)^\top + \eta^2 (\nabla_{a_r} L) (\nabla_{w_r} L)^\top  \right) x_i\1[ w_r(k)^\top x_i \geq 0],
\end{align*}
where the first part corresponds to kernel $G(t)$ and the second part corresponds to the $H(t)$ shown up in the gradient flow analysis. With this intuition, we split the right hand side into two parts. $v_{1,i}$ represents the terms that the pattern does not change and $v_{2,i}$ represents the remaining term that pattern may changes.

For each $i \in [n]$, we define $S_i = \{r \in [m] : \1[ w_r(k+1)^\top x_i \geq 0] = \1[ w_r(k)^\top x_i \geq 0 ]$, and $\Sc_i = [m] \backslash S_i$.
Then, we write  $v_{1,i}$ and $v_{2,i}$ as follows:
\begin{align*}
&v_{1,i} \triangleq \frac{1}{ \sqrt{md} } \sum_{r \in S_i}  \left( a_r(k+1) \phi( w_r(k+1)^\top x_i ) - a_r(k)\phi(w_r(k)^\top x_i ) \right), \\
&v_{2,i} \triangleq \frac{1}{ \sqrt{md} } \sum_{r \in \Sc_i}  \left( a_r(k+1) \phi( w_r(k+1)^\top x_i ) - a_r(k)\phi(w_r(k)^\top x_i ) \right) .
\end{align*}
We further write $v_1 = (v_{1,1}^\top, v_{1,2}^\top, \dots, v_{1,n}^\top)^\top$ and do the same for $v_2$. Hence, we write 
\begin{align*}
\vec(U(k+1) - U(k)) = v_1 + v_2 .
\end{align*}

In order to analyze $v_1 \in \R^n$, we provide definition of $G, G^{\bot} \in \R^{nd \times nd}$ and $H, H^{\bot} \in \R^{nd \times nd}$,
\begin{align*}
  G(k)_{i,j} = & ~ \frac{1}{m} \sum_{r=1}^m x_i^\top x_j \1[{ w_r(k)^\top x_i \geq 0, w_r(k)^\top x_j \geq 0 }] a_r(t)a_r(t)^\top, \\
  G(k)^{\bot}_{i,j} = & ~ \frac{1}{m} \sum_{r\in \Sc_i} x_i^\top x_j \1[ w_r(k)^\top x_i \geq 0, w_r(k)^\top x_j \geq 0]a_r(t)a_r(t)^\top, \\
  H(k)_{i,j} = & ~ \frac{1}{m} \sum_{r=1}^m \phi(w_r(t)^\top x_i) \phi(w_r(t)^\top x_j)I, \\
  H(k)^{\bot}_{i,j} = & ~ \frac{1}{m} \sum_{r\in \Sc_i} \phi(w_r(t)^\top x_i) \phi(w_r(t)^\top x_j)I.
\end{align*}
Using the fact that $\phi(z) = z\1[z \geq 0]$ and the definition of $S_i$, we expand the forms of the gradients in \ref{eqnGradLossOverWr_joint} and \ref{eqnGradLossOverAr_joint} and get:
\begin{align*}
  v_{1,i}
  = &~- \frac{1}{\sqrt{md}} \sum_{r \in S_i} \eta a_r(k) \left( \nabla_{w_r} L  \right)^\top x_i\1[ w_r(k)^\top x_i \geq 0] \\
  &~- \frac{1}{\sqrt{md}} \sum_{r \in S_i} \eta ( \nabla_{a_r} L ) w_r(k)^\top x_i\1[ w_r(k)^\top x_i \geq 0] \\
    &~+ \frac{1}{\sqrt{md}} \sum_{r \in S_i} \eta^2 (\nabla_{a_r} L) (\nabla_{w_r} L)^\top x_i\1[ w_r(k)^\top x_i \geq 0] \\
  = &~ \frac{\eta}{d} \sum_{j=1}^n \left( G(k)_{i, j} - G(k)_{i, j}^\bot + H(k)_{i, j} - H(k)_{i, j}^\bot \right) (x_j - u_j) + v_{3, i},
\end{align*}
where $v_3$ will be treated as a perturbation:
\[
  v_{3, i} = \frac{\eta^2}{\sqrt{md}} \sum_{r \in S_i} (\nabla_{a_r} L) (\nabla_{w_r} L)^\top x_i\1[ w_r(k)^\top x_i \geq 0].
\]
Then, we can write $v_1$ as:
\begin{align}
  \label{eqnV1compact}
  v_1 = \frac{\eta}{d} (K(k) - K^{\bot}(k)) \vec(X - U(k))  + v_3,
\end{align}
in which $K(k) = G(k) + H(k)$ --- the discrete NTK kernel and $K^\bot (k) =  H^\bot(k) + G^\bot(k)$. Lastly, we come to the main prediction dynamics in discrete time for $\vec(U(k+1) - U(k))$ as:
\begin{align*}
\vec(U(k+1) - U(k)) = \frac{\eta}{d} (K(k) - K^{\bot}(k)) \vec(X - U(k) + v_2 + v_3.
\end{align*}
Using this equation, we can rewrite $\| X - U(k+1) \|_2^2$ in terms of $X - U(k)$ as follows:
\begin{align*}
  \norm{ X - U(k+1)}_F^2
  &= \norm{ \vec(X - U(k+1))}^2 \\
  &= \norm{ \vec(X - U(k)) - \vec(U(k+1) - U(k)) }_F^2 \\
  &= \norm{X - U(k)}_F^2 - 2\vec(X - U(k))^\top \vec(U(k+1) - U(k)) \\
  &~~~+ \norm{U(k+1) - U(k)}_F^2  \\
  &= \norm{X - U(k)}_F^2 - \frac{2\eta}{d}\vec(X - U(k))^\top K(k) \vec(X - U(k)  \\
  &~~~+ \frac{2\eta}{d}\vec(X - U(k))^\top K(k)^\bot \vec(X - U(k) \\
  &~~~- \frac{2\eta}{d}\vec(X - U(k))^\top (v_2 + v_3) \\
  &~~~+ \norm{U(k+1) - U(k)}_F^2.
\end{align*}
We define and upper bound each of the following terms
\begin{align*}
  C_1 &= -\frac{2\eta}{d} \vec( X - U(k) )^\top K(k) \vec( X - U (k) ) , \\
  C_2 &= \frac{2\eta}{d} \vec( X - U(k) )^\top K(k)^\bot \vec( X - U (k) ) , \\
  C_3 &= -\frac{2\eta}{d} \vec( X - U(k) )^\top v_2 , \\
  C_4 &= -\frac{2\eta}{d} \vec( X - U(k) )^\top v_3  , \\
  C_5 &= \norm{U(k+1) - U(k)}_F^2 . 
\end{align*}
Notice that $C_1$ can be upper bounded in terms of $\lambda_{\min}(K(k)) \ge \lambda_{\min}(H(k))$, which is ensured as long as the movement in the weight is sufficiently small (shown in Lemma \ref{lmMovementInstep_joint}.) $C_2$ can be upper bounded also using the kernel with bound on its spectral norm.

\proof[Proof of Theorem \ref{thmGradDescent_joint}]
We will prove Theorem \ref{thmGradDescent_joint} by induction. The base case when $k = 0$ is trivially true. Assume that the claim holds for $k'=0, 1, \ldots, k$ and we want to show that \eqref{eqnGDConvergence_joint} also holds for $k' = k+1$. For $k' = k+1$, we have
\begin{align*}
\norm{ X - U(k+1)}_F^2
&= \norm{X - U(k)}_F^2 + C_1 + C_2 + C_3 + C_4 + C_5
\end{align*}
Now, we invoke the bound for each of these terms from Claims~\ref{clmC1_joint}, \ref{clmC2_joint}, \ref{clmC3_joint}, \ref{clmC4_joint} and \ref{clmC5_joint} in Appendix \ref{appdx:jointly_train} and Lemma\ref{lmMovementInstep_joint}. Then, we want to choose $\eta$ and $R_w$ such that
\begin{align}\label{eq:choice_of_eta_R}
\left( 1 - \frac{\eta \lambda_0}{d} + \frac{8\eta \lambda_n}{d} +  \frac{16\eta^2 \sqrt{n\lambda_n}}{d}  + \frac{8\eta^2 \sqrt{n\lambda_n^2}}{d} \norm{X - U(0)}_F + \frac{64 \eta^2  \lambda_n^2}{md}  \right) \leq \left(1-\frac{\eta\lambda_0}{2d} \right) .
\end{align}

If we set $\eta = \frac{\lambda_0 }{64n\lambda_n}$ and use $\norm{X - U(0)}_F \leq C\sqrt{n}$, we have the two dominating terms are
\begin{align*}
\frac{8\eta\lambda_n}{d} \leq  \frac{\eta \lambda_0 }{8d} ,
\mathrm{~~~and~~~} \frac{8\eta^2 \sqrt{n\lambda_n^2}\norm{X - U(0)}_F}{d} \leq  \frac{\eta \lambda_0}{8d}.
\end{align*}
This implies that
\begin{align*}
\norm{ X - U(k+1) }_F^2 \leq ( 1 - \frac{\eta \lambda_0}{2d} ) \norm{ X - U(k) }_F^2.
\end{align*}

\paragraph{Lower bound on $m$.}
We require for any $\delta \in (0, 1)$ that
\begin{align*}
\frac{\lambda_n}{m}(4\norm{W(0)} + R_w)R_w \leq \frac{\lambda_0}{4},
\end{align*}
\begin{align*}
  R_w \leq \frac{4\sqrt{d\lambda_n}(\norm{A(0)} + R_a) \norm{X - U(0)}_F}{\sqrt{m}\lambda_0}
\end{align*}
and
\[
  2\exp(-m) \leq \delta
\]
where the first bound on $R_w$ comes from the result on gradient descent and the condition in Lemma \ref{lmBoundHDiff_joint}, whereas the second bound is required by the above Claims. By Claim \ref{clInitialLoss_Case1} that $\norm{X - U(0)}_F \leq \sqrt{\frac{2n}{\delta}}$ for arbitrary $\delta \in (0, 1)$, then we require
\[
  m \geq C\frac{nd\lambda_n^3}{\lambda^4_0\delta^2}
\]
for a sufficiently large constant $C>0$ so that the claim holds with probability $1-\delta$.

\qedhere


\section{Weight-tied Autoencoders}
\label{sec:scaling}

We conclude with the case of training two-layer autoencoders whose weights are shared (i.e., $A = W$). This is a common architectural choice in practice, and indeed previous theoretical analysis for autoencoders~\citep{tran17,nguyen2019_ae, li2018_randomAE} have focused on this setting. We will show that somewhat surprisingly, allowing the network to be over-parameterized in this setting leads to certain degeneracies. First, we prove:

\begin{Lemma}
  \label{lmExpectedInitLoss_shared}
Let $x$ be any fixed sample. The weight $W$ is randomly initialized such that $w_r \sim \normal(0, \sigma^2I)$ independently for $r=1, 2, \dots, m$, then
\[
\E_{w_r \sim \normal(0, \sigma^2I), \forall r}[\norm{x - \frac{1}{m}W\phi(W^\top x)}^2] = \left(\frac{\sigma^2}{2} - 1\right)^2 \norm{x}^2 + \frac{(2d+3)\norm{\sigma^2 x}^2}{4m}.
\]
Particularly, when $\norm{x}=1$, $\sigma^2=2$, then
\[
  \E_{w_r \sim \normal(0, \sigma^2I), \forall r}[\norm{x - \frac{1}{m}W\phi(W^\top x)}^2] = \frac{2d+3}{m}.
\]
For an arbitrary small $\epsilon > 0$, the expected reconstruction loss is at most $\epsilon$ if $m \ge \Omega(d/\epsilon)$. 
\end{Lemma}
\begin{Remark}
  \normalfont  This Lemma has a few interesting implications. First, when $\sigma^2 = 2$ 
  , then
\[
  \E_{w_r \sim \normal(0, 2I), \forall r}[\norm{x - u}^2] = \frac{(2d+3)\norm{x}^2}{m},
\]
which does not exceed $\epsilon$ if $m \ge 3d/\epsilon$ for $\epsilon > 0$. Provided that the data samples are normalized, if $m$ is sufficiently large, even with random initialization the reconstruction loss is very close to zero \emph{without any need for training}. Therefore, mere over-parameterization already gets us to near-zero loss; the autoencoder mapping $\frac{1}{m} W\phi(W^\top x) \approx x$ for any unit-norm $x$. It suggests that training of weight-tied autoencoders under high levels of over-parameterization may be degenerated.
\end{Remark}

\proof We will use $\E_{w_r}$ as a shorthand for $\E_{w_r \sim \mathcal{N}(0, \sigma^2I)}$. We expand the reconstruction loss:
\begin{align}
\label{eqn:loss_at_W0}
\norm{x - \frac{1}{m}W\phi(W^\top x)}^2 &= \norm{x - \frac{1}{m}W\phi(W^\top x)}^2 = \norm{x - \frac{1}{m}\sum_{r=1}^m \phi(w_r^\top x)w_r}^2 \nonumber \\
&= \norm{x}^2 - \frac{2}{m}\sum_{r=1}^m x^T\phi(w_r^\top x)w_r  + \frac{1}{m^2}\sum_{r, s \in [m]} \phi(w_r^\top x)w_r^\top \phi(w_s^\top x)w_s.
\end{align}
Because $\phi$ is ReLU and the distribution of $w_r$ is symmetric, we have:
\begin{align*}
  \E_{w}[\phi(\wT x)w] = \frac{1}{2} \E_w[w\wT]x = \frac{\sigma^2x}{2},
\end{align*}
since $\E_w[w\wT]= I$. Then, by the independence of the columns in $W$ (more details or split up, one more step),
\begin{align}
\label{eqn:expected_loss1}
\E_{w_r}\bigl[ \norm{x - u}^2 \bigr] &= \norm{x}^2 - \frac{2}{m}\sum_{r=1}^m \frac{\norm{x}^2}{2} + \frac{1}{m^2}\sum_{r, s\in [m], r \neq s} \frac{\norm{\sigma^2 x}^2}{4} + m\E_w[\phi(w^Tx)^2\norm{w}^2] \nonumber \\
&= (1-\sigma^2)\norm{x}^2 + \frac{m-1}{4m}\norm{\sigma^2 x}^2 + \frac{1}{m}\E_w[\phi(w^Tx)^2\norm{w}^2].
\end{align}
Now, we compute the last term:
\begin{align}
  \E_w[\phi(w^Tx)^2\norm{w}^2] = \frac{d+2}{2}\norm{\sigma^2 x}^2.
  \label{eqn:fact2}
\end{align}

Due to the normalization $\|x\|=1$, we can also write $ \frac{1}{\sigma}w = ux + v$ such that $x^Tv = 0$, then $u = \inprod{w}{x} \sim \mathcal{N}(0, 1)$ and $v \sim \mathcal{N}(0, I-xx^T)$ are conditionally independent given $x$. Note that since the conditional distribution of $u$ is unchanged with respect to $x$, this implies that $u$ is independent of $x$; as a result, $u$ and $v$ are (unconditionally) independent.

Also, denote $\alpha_q = \E_{z \sim \mathcal{N}(0, 1)}[z^q\1(z \ge 0)]$ 
for the exact value
. Using Stein's Lemma, we can compute the exact values: $\alpha = \E_w[u\1(u \ge 0)] =  \E_{z \sim \mathcal{N}(0, 1)}[z\1(z \ge 0)] = \frac{1}{\sqrt{2\pi}}$, $\beta = \E_w[u^2\1(u \ge 0)] = \frac{1}{2}$, $\gamma = \E_w[u^4\1(u \ge 0)] = \frac{3}{2}$, which are all positive.
Write $\phi(z) = \max(0, z) = \1(z \ge 0)z$, and 
\begin{align}
\label{eq_square_term}
 \E_{w \sim \mathcal{N}(0, I)}[\phi(w^Tx)^2\norm{w}^2]
 &=\E_w[\1(\inprod{w}{x} \ge 0)\inprod{w}{x}^2\norm{w}^2] \nonumber \\
 &= \E_w[\1(u \ge 0) u^2 (u^2 + \norm{v}^2)] ~~(\norm{x} = 1) \nonumber \\
 &= \E_w[\1(u \ge 0) u^4] + \E_w[\1(u \ge 0)u^2] \E_w[\norm{v}^2] ~~\text{(cond. independence of $u$ and $v$.)} \nonumber \\
 &= \frac{d + 2}{2}.
\end{align}
Changing variables by scaling the variance:
\begin{align*}
  \E_{w \sim \mathcal{N}(0, \sigma^2 I)}[\phi(w^Tx)^2\norm{w}^2] &=  \sigma^4\E_{w \sim \mathcal{N}(0, I)}[\phi(w^Tx)^2\norm{w}^2] = (2d + 4)\norm{\sigma^2 x}^2 .
\end{align*}
Combining with~\eqref{eqn:expected_loss1}:
\begin{align*}
\label{eqn:expected_loss2}
  \E_{w_r \sim \mathcal{N}(0, \sigma^2I)}\bigl[ \norm{x - u}^2 \bigr] &= \left(\frac{\sigma^2}{2} - 1\right)^2 \norm{x}^2 + \frac{(2d+3)\norm{\sigma^2 x}^2}{4m} .
  \numberthis
\end{align*}
The second result directly follows from the Lemma with the specific values of $\norm{x}, \sigma$ plugged in.

\qedhere



\clearpage
\bibliographystyle{unsrtnat}
\bibliography{refs}

\appendix
\section{Useful Facts}
\label{apdx:facts}

\begin{Lemma}[Stein's Lemma]
\label{lm:stein}
For a random vector $w \in R^{d}$ such that $w \sim \normal(0, I)$ and function $h(w) : \R^d \rightarrow \R^k$ is weakly differentiable with Jacobian $D_w h$, we have
\[
\E_{w \sim \normal(0, I)}\bigl[wh(w)^\top\bigr] = \E_{w \sim \normal(0, I)}\bigl[(D_w h)^\top\bigr].
\]
\end{Lemma}



\begin{Lemma}[]
  \label{lmSongClaim4.10}
  Denote $S_i = \{r \in [m] : \1[ w_r(k+1)^\top x_i \geq 0] = \1[ w_r(k)^\top x_i \geq 0 ]$, and $\Sc_i = [m] \backslash S_i$. If $\norm{w_r(k) - w_r(0)} \leq R$, then
  \[
    \sum_{r=1}^m \1[r \in \Sc_i] \leq 4mR
  \]
  with probability at least $1 - n\exp(-mR)$.
\end{Lemma}
This result is borrowed from the proof of \cite[Claim 4.10]{song2019quadratic}.



\section{Weakly-trained Autoencoders}
\label{appdx:weakly_train}

\subsection{Concentration of $K(0)$ -- Proof of Lemma \ref{lmMinEigenK0_Case1}}
\proof Recall that $K(0) = \frac{1}{m}\sum_{r=1}^m \Xt_r(0)^\top \Xt_r(0) \otimes a_r\aT_r$.
In this proof, we omit the argument $t=0$ in $\Xt_r(0)$, and simply write $\Xt_r$ for clarity.

Consider the random matrix $Z_r = \Xt_r^\top \Xt_r \otimes a_r\aT_r$ and $\bar{Z}_r = \E_{w_r}[\Xt^\top \Xt] \otimes I$. Note that $Z_r$ is positive semi-definite.
One can easily show two facts:
\[
  \norm{Z_r}=\norm{\Xt_r^\top \Xt_r \otimes a_r\aT_r}  = \norm{\Xt_r^\top \Xt_r}\norm{ a_r\aT_r} = \norm{a_r}^2\norm{\Xt_r^\top \Xt_r} \leq d\lambda_n,
\]
in which we use $\norm{a_r}^2=d$; and
\begin{equation}
  \norm{\Xt_r^\top\Xt_r} = \sup_{\norm{b} = 1}\norm{\Xt_r b}^2 \leq \sup_{\|b\|=1} \|\sum_{i} b_i x_i\|^2 = \norm{X^\top X} = \lambda_n.
  \label{eqn:tXnormbound}
\end{equation}
Similarly, $\norm{\bar{Z}_r}\le \lambda_n$, and hence $\norm{Z_r-\bar{Z}_r}\le (d+1)\lambda_n$. Moreover,
\begin{align*}
  \E_{w_r, a_r}[(Z_r-\bar{Z}_r)^2]
  &=\E_{w_r, a_r}[(Z_r-\bar{Z}_r)(Z_r - \bar{Z}_r)^\top]\\
  &= \E_{w_r,a_r}[Z_rZ_r^\top] - \bar{Z}_r^2 \\
  &= \E_{w_r, a_r}[(\Xt_r^\top \Xt_r)^2 \otimes \norm{a_r}^2a_r\aT_r] - (\E_w[\Xt^\top \Xt])^2 \otimes I \\
  &\preceq d\E_{w_r}[(\Xt_r^\top \Xt_r)^2] \otimes I.
\end{align*}
By the above argument, $\norm{\E_{w_r}[(\Xt_r^\top \Xt_r)^2]}\le \lambda_n^2$,
so $\|\sum_r \E((Z_r-\bar{Z}_r)^2)\|\le md\lambda_n^2$.

From matrix Bernstein's inequality \citep[Theorem 1.4 of]{Tropp12},
\begin{align*}
  \Prob\left[ \norm{mK(0) - mK^\infty} \geq \epsilon \right] \leq nd \exp\left(- \frac{\epsilon^2/2}{(d+1)\lambda_n \epsilon/3 + md\lambda_n^2} \right).
\end{align*}
Since the second term in the denominator of the exponent dominates ($\lambda_0 \leq \lambda_n$), we get
\[
  m \ge C
  \frac{\lambda^2_nd\log(nd/\delta)}{\lambda^2_0}
\]
where we pick $\epsilon = m\lambda_0/4$. Therefore,
\[
  \norm{K(0) - K^\infty} \leq {\lambda_0}/{4}
\]
with probability at least $1 -\delta$ for any $\delta \in (0, 1)$. By Weyl's inequality, we have with the same probability: 
\[
  \lambda_{\min}(K(0)) \geq 3\lambda_0/4.
\]
\qedhere

\subsection{Proof of supporting claims}

To prove the bounds in Claims \ref{clmC1_Case1}, \ref{clmC2_Case1}, \ref{clmC3_Case1}, and \ref{clmC4_Case1}, we use the bound $\norm{w_r(k+1) - w_r(0)} \leq R'$ for all $r \in [m]$ in Lemma \ref{lmMovementInstep_Case1}. In what follows, we assume $R' < R$, which is the weight movement allowed to achieve Lemma \ref{lmBoundKDiff_Case1}. This assumption holds with high probility as long as $m$ is large enough.

\begin{Claim}
  \label{clmC1_Case1}
  Let $C_1 = -\frac{2\eta}{d} \vec( X - U(k) )^\top K(k) \vec( X - U (k) )$. Then we have
  \begin{align*}
    C_1 \leq -\frac{\eta\lambda_0}{d} \norm{X - U(k)}_F^2.
  \end{align*}
  with probability at least $1-\delta$.
\end{Claim}
\proof
Using Lemma \ref{lmMovementInstep_Case1}, we have $\norm{w_r(k) - w_r(0)} \leq R' < R$ for all $r \in [m]$. By Lemma \ref{lmBoundKDiff_Case1},
we have
\[
  \|K(k)-K(0)\| < \frac {\lambda_0}{4}.
\]
Therefore, $\lambda_{\min}(K(k)) \ge \lambda_0 / 2$ with probability at least $1-\delta$. As a result,
\begin{align*}
  \vec(X - U(k))^\top K(k) \vec( X - U(k) ) \geq  \frac{\lambda_0}{2} \| X - U(k) \|^2= \frac{\lambda_0}{2} \norm{X - U(k) }_F^2,
\end{align*}
and $C_1 \leq -\frac{\eta\lambda_0}{d} \norm{X - U(k)}_F^2$ with probability at least $1-\delta$.

\qedhere

\begin{Claim}
  \label{clmC2_Case1}
  Let $C_2 =  \frac{2\eta}{d} \vec( X - U(k) )^\top K(k)^{\bot} ( X - U(k) )$. We have
  \begin{align*}
    C_2 \leq 8\eta nR \norm{X - U(k)}_F^2.
  \end{align*}
  with probability at least $1-n\exp(-mR)$.
\end{Claim}
\proof
All we need is to bound $K(k)^{\bot}$. A simple upper bound is
\begin{align*}
  \norm{K(k)^{\bot}}^2 &\leq \sum_{i, j=1}^n \norm{K(k)^{\bot}_{i, j}}^2_F \\
                     &\leq \sum_{i, j=1}^n \norm[\Big]{ \frac{1}{m} \sum_{r\in \Sc_i} x_i^\top x_j \1[ w_r(k)^\top x_i \geq 0, w_r(k)^\top x_j \geq 0]a_r\aT_r }^2_F \\
                       &\leq d^2\sum_{i, j=1}^n \left( \frac{1}{m} \sum_{r\in \Sc_i} \1[ w_r(k)^\top x_i \geq 0, w_r(k)^\top x_j \geq 0] \right)^2 \\
                       &\leq d^2\sum_{i, j=1}^n \left( \frac{1}{m} \sum_{r=1}^m \1[r\in \Sc_i] \right)^2 \\
                       &\leq 16n^2d^2R^2
\end{align*}
with probability $1-n\exp(-mR)$ where the last step follows from Lemma \ref{lmSongClaim4.10}. Then, with that same probability $\norm{K(k)^{\bot}} \leq \norm{K(k)^{\bot}}_F \leq 4ndR$, and 
\begin{align*}
  C_2 &=  \frac{2\eta}{d} \vec( X - U(k) )^\top K(k)^{\bot} ( X - U(k) ) \\
                         &\leq \frac{2\eta}{d} \norm{K(k)^{\bot}} \norm{X - U(k)}_F^2 \\
                         &\leq 8\eta nR \norm{X - U(k)}_F^2.
\end{align*}

\qedhere

\begin{Claim}
  \label{clmC3_Case1}
  Let $C_3 = -2\vec( X - U(k) )^\top v_2$, then   with  probability at least $1 - n\exp(-mR)$
  \begin{equation*}
    C_3 \leq 8\eta n R \norm{X - U(k)}_F^2.
  \end{equation*}
\end{Claim}
\proof
We have $C_3 \leq 2\norm{ X - U(k) }_F \norm{v_2}$. Using the Lipschitz property of $\phi$, we have
\begin{align*}
  \norm{v_2}^2 &= \sum_{i=1}^n \norm{v_{2, i}}^2 \\
               &\leq \sum_{i=1}^n \norm[\bigg]{\frac{1}{ \sqrt{md} } \sum_{r \in \Sc_i} a_r \left( \phi \left( w_r(k+1)^\top x_i \right) - \phi( w_r(k)^\top x_i ) \right) }^2 \\
               &\leq \frac{\eta^2}{ m } \sum_{i=1}^n \left( \sum_{r \in \Sc_i} \left|\Big( \nabla_{w_r}L(W(k)) \Big)^\top x_i\right| \right)^2 \\
               &\leq \frac{\eta^2}{ m } \sum_{i=1}^n \left( \sum_{r =1}^m \1[r \in \Sc_i]  \left|\Big( \nabla_{w_r}L(W(k)) \Big)^\top x_i\right| \right)^2 \\
               &\leq \frac{\eta^2}{ m }\max_r \norm[\Big]{ \nabla_{w_r}L(W(k)) \Big)^\top }^2 \sum_{i=1}^n\left( \sum_{r=1}^m \1[r \in \Sc_i] \right)^2 \\
               &\leq \frac{\eta^2\lambda_n}{ m^2 }\norm{X - U(k)}_F^2 \sum_{i=1}^n\left( \sum_{r=1}^m \1[r \in \Sc_i] \right)^2 \\
               &\leq \frac{\eta^2\lambda_n}{ m^2 }\norm{X - U(k)}_F^2 \sum_{i=1}^n (4mR)^2 \\
               &\leq 16n\lambda_nR^2\eta^2 \norm{X - U(k)}_F^2 \\
               &\leq 16n^2R^2\eta^2 \norm{X - U(k)}_F^2.
\end{align*}
with probability $1 - n\exp(-mR)$. The sixth step we use
\begin{align*}
  \norm{\nabla_{w_r}L(W(k))} &= \norm[\Big]{\frac{1}{\sqrt{md}}\Xt_r(k)(X - U(k))^\top a_r} \\
                             &\leq \frac{\sqrt{\lambda_n}}{\sqrt{m}}\norm{ X - U(k) }_F,
\end{align*}
and the last step follows from from Lemma \ref{lmSongClaim4.10} that $\sum_{r=1}^m \1[r \in \Sc_i] \leq 4mR$ with  probability at least $1 - n\exp(-mR)$. Substitute the bound into $C_3$, then we finish the proof.
\qedhere

\begin{Claim}
  \label{clmC4_Case1}
  Let $C_4  = \norm{ U (k+1) - U(k) }_F^2$. Then we have
  \begin{align*}
    C_4 \leq \eta^2 n\lambda_n \| X - U(k) \|_F^2.
  \end{align*}
\end{Claim}
\proof
Previously in Lemma \ref{lmMovementInstep_Case1}, we proved that
\begin{align*}
  \norm{\nabla_{w_r}L(W(k))} &= \norm[\Big]{\frac{1}{\sqrt{md}}\Xt_r(k)(X - U(k))^\top a_r} \\
                             &\leq \frac{\sqrt{\lambda_n}}{\sqrt{m}}\norm{ X - U(k) }_F.
\end{align*}
Expand the form of $U(k+1) - U(k)$ and use the Lipschitz of $\relu$ to get
\begin{align*}
  C4 &= \sum_{i=1}^n \norm{ u_i(k+1) - u_i(k) }^2 \\
     &= \frac{1}{md} \sum_{i=1}^n \norm[\Big]{ \sum_{r=1}^m a_r \left( \phi( w_r(k+1)^\top x_i ) - \phi(w_r(k)^\top x_i ) \right)}^2 \\
     &\leq \eta^2 \sum_{i=1}^n \frac{1}{m} \left( \sum_{r=1}^m \norm[\Big]{\nabla_{w_r}L( W(k) ) } \right)^2 \\
     &\leq \eta^2 \sum_{i=1}^n \frac{1}{m} \left( \sum_{r=1}^m \frac{\sqrt{\lambda_n}}{\sqrt{m}}\norm{ X - U(k) }_F \right)^2 \\
     &= \eta^2 n\lambda_n \norm{ X - U(k) }_F^ 2 \\
     &\leq n^2\eta^2 \norm{ X - U(k) }_F^2.
\end{align*}
Therefore, we finish the proof.
\qedhere

\section{Jointly-trained Autoencoders}
\label{appdx:jointly_train}

\subsection{Concentration of $H(0)$}
We re-state and prove the concentration of $H(0)$ in Lemma \ref{lmMinEigenH0_joint}.
\begin{Lemma}
  For any $\delta \in (0, \frac{\lambda_0}{12d\lambda_n^2})$, if $m \ge C\frac{\max(n, d) \lambda_n\log^2(nd/\delta)}{\lambda_0^2}$ for some large enough constant $C$, then with probability at least $1 - 1/(2nd)^{2\log nd} - m\delta$,
  one obtains $\norm{H(0) - H^\infty} \leq {\lambda_0}/{4}$ and $\lambda_{\min}(H(0)) \geq 3\lambda_0/4$ .
\end{Lemma}
\proof
Recall that
\begin{align}
  H(0) = \frac{1}{m}\sum_{r=1}^m\phi(X^\top w_r(0))\phi(w_r(0)^\top X) \otimes I,
\end{align}
and $H^\infty = \E_{W(0), A(0)}[H(0)]$. Our goal is to show the concentration of $\sum_{r=1}^m\phi(X^\top w_r(0))\phi(w_r(0)^\top X)$. Let us use $w_r$ to mean $w_r(0)$ and denote the $r^{\textrm{th}}$-random matrix as
\[
  Z_r = \phi(X^\top w_r)\phi(w_r^\top X).
\]
We use Lemma B.7 of \citet{zhong2017recovery} and
verify the required conditions by
the next results. 

\begin{Claim}[Condition I for H(0)] The following are true:
  \begin{enumerate}[label=(\roman*).,leftmargin=*]
  \item $\norm{Z_r} \leq \lambda_n\norm{w_r}^2.$
  \item $\Prob_{w_r}[\norm{w_r}^2 \leq 4d\sqrt{\log(2/\delta)}] \geq 1 - \delta$ for any $\delta \in (0, 1) $.
  \end{enumerate}
  \label{clCond1_H0_joint}
\end{Claim}
\proof
(i) We have
\[
  \norm{Z_r} = \norm{\phi(X^\top w_r)\phi(\wT_r X)} = \norm{\phi(X^\top w_r)}^2 \leq \norm{X^\top X} \norm{w_r}^2 \leq \lambda_n\norm{w_r}^2,
\]
which gives the first part (i). For the second part, we use the fact $\norm{w_r}^2$ is a chi-squared random variable with $d$ degrees of freedom, and sub-exponential with sub-exponential norm $2\sqrt{d}$, meaning that:
\[
  \Prob_{w_r}[|\norm{w_r}^2 - d| \geq \epsilon] \leq 2\exp\bigl(-\frac{\epsilon^2}{8d}\bigr)
\]
For any $\delta \in (0, 1)$ and $\epsilon^2 = 9d\log(2/\delta)\geq 8d\log(2/\delta)$, we have
\[
  |\norm{w_r}^2 - d| \leq  3\sqrt{d\log(2/\delta)}
\]
with probability at least $1-\delta$. Then, $\norm{w_r}^2 \leq d + 3\sqrt{d\log(2/\delta)} \leq 4d\sqrt{\log(2/\delta)}$ with probability at least $1-\delta$.
\qedhere

\begin{Claim}[Condition II for H(0)]
  $\norm{\E[Z_rZ_r^\top]} \leq 3n\lambda_n.$
  \label{clCond2_H0_joint}
\end{Claim}
\proof  We have
\begin{align*}
  Z_rZ_r^\top &= \phi(X^\top w_r)\phi(\wT_r X)\phi(X^\top w_r)\phi(\wT_r X)\\
             &\preceq \norm{\phi(X^Tw_r)}^2 \norm{w_r}^2 \XT X \preceq  \sum_{l=1}^n \phi(\wT x_l)^2\norm{w_r}^2 \XT X.
\end{align*}
We need is to compute $\E_{w_r}[\phi(\wT x_l)^2\norm{w_r}^2]$, which is already done in \eqref{eqn:fact2}, Section \ref{sec:scaling}. To be precise, we have
\begin{align*}
  \E[\phi(\wT_r x_l)^2\norm{w_r}^2] = \frac{d+2}{2} \le d
\end{align*}
where $\xt_i = x_i1[\wT_r x_i \geq 0]$ for each $i \in n$. Then, we can write
\[
  \norm{\E_{w_r}[Z_rZ_r^\top]} \leq nd\lambda_n.
\]
\qedhere

\begin{Claim}[Condition IV for H(0)]
  $\sup_{\{b:\norm{b}=1\}}(\E[(b^\top Z_r b)^2])^{1/2} \leq \sqrt{3}d\lambda_n$.
  \label{clCond3_H0_joint}
\end{Claim}
\proof Recall $Z_r = \phi(X^\top w_r)\phi(w_r^\top X)$, and for any unit-norm vector $b \in \R^n$
\[
  (b^\top Z_r b)^2 = \norm{b^\top \phi(X^\top w_r)}^4 \leq \norm{\phi(X^\top w_r)}^4 \leq \lambda_n^2 \norm{w_r}^4.
\]
Moreover, $\norm{w_r}^2$ is a chi-squared random variable with $d$ degree of freedom, so 
\[
  \E[\norm{w_r}^4] = 3d^2.
\]
Therefore,   $\sup_{\{b:\norm{b}=1\}}(\E[(b^\top Z_r^\top b)^2])^{1/2} \leq \sqrt{3}d\lambda_n$.
\qedhere

\proof[Proof of Lemma \ref{lmMinEigenH0_joint}.] With the conditions fulfilled in Claims \ref{clCond1_H0_joint}, \ref{clCond2_H0_joint} and \ref{clCond3_H0_joint}, we can now apply \cite[Lemma B.7]{zhong2017recovery} to show the concentration of $K(0)$:
\[
  \norm[\big]{\frac{1}{m}\sum_{r=1}^mZ_r - \E[Z_r]} \leq \epsilon \norm{\E[Z_r]}
\]
with probability $1-1/n^{2t}-n\delta$ for any $t \geq 1$, $\epsilon \in (0, 1)$ and $\delta < \epsilon \norm{\E[Z_r]}/(2\sqrt{3}d\lambda_n)^2$.

For the target bound, we choose $\epsilon \norm{\E[Z_r]} = \lambda_0/4$, $t = \log (2nd)$ and note that $\lambda_0 \leq \norm{\E[Z_r]} = \norm{K^\infty} \leq \lambda_n$. Therefore, with probability $1- 1/(2nd)^{2\log nd} - m\delta$ for any $\delta \in (0, \frac{\lambda_0}{12d^2\lambda_n^2})$, then
\[
  \norm{H(0) - H^{\infty}} \leq \frac{\lambda_0}{4}
\]
if $m$ satisfies
\[
  m \geq 18\log^2 (2nd)\frac{nd\lambda_n + \lambda_n^2 + (4d\sqrt{\log(2/\delta)})\lambda_n\lambda_0/4}{\lambda_0^2},
\]
which means $ m =  C\frac{nd \lambda_n \log^2(nd)\log(1/\delta)}{\lambda_0^2}$.
\qedhere

\subsection{Proof of supporting claims}
In the proof of the next claims, we assume that $\norm{W(0)} \geq R'_w > R_w$ and $\norm{A(0)} \ge R'_a > R_a$. Also, assume $d \ll m$. These conditions will hold with high probability when $m$ is large enough.

\begin{Claim}
  \label{clmC1_joint}
  Let $C_1 = -\frac{2\eta}{d} \vec( X - U(k) )^\top K(k) \vec( X - U (k) )$ . We have
  \begin{align*}
    C_1 \leq -\frac{\eta\lambda_0}{d} \norm{X - U(k)}_F^2.
  \end{align*}
\end{Claim}
\proof
Since we have proved that $\| W(k) - W(0) \|_F \leq R_w'$, using Lemma \ref{lmBoundHDiff_joint} with the choice of $R_w  < R_w'$,
we have
\[
  \|H(k)-H(0)\| \leq \frac {\lambda_0}{4}.
\]
Moreover, $G(k)$ is p.s.d, therefore $\lambda_{\min}(K(k)) \geq \lambda_{\min}(H(k)) \geq \lambda_0 / 2$, and as a result,
\begin{align*}
  \vec(X - U(k))^\top K(k) \vec( X - U(k) ) \geq  \frac{\lambda_0}{2} \| X - U(k) \|^2= \frac{\lambda_0}{2} \norm{X - U(k) }_F^2.
\end{align*}
and $    C_1 \leq -\frac{\eta\lambda_0}{d} \norm{X - U(k)}_F^2$.
\qedhere

\begin{Claim}
  \label{clmC2_joint}
  Let $C_2 = \frac{2\eta}{d} \vec( X - U(k) )^\top K(k)^\bot \vec( X - U (k) ) $. We have
  \[
    C_2 \leq \frac{8\eta \lambda_n}{d} \norm{X - U(k)}_F^2
  \]
  with probability at least $1-2\exp(-m)$.
\end{Claim}
\proof
We need to bound the spectral norm of $K(k)^\bot$, defined as $K(k)^\bot = G(k)^\bot + H(k)^\bot$. We will bound their spectral norms. We have
\begin{align*}
  \norm{G(k)^\bot} &= \norm[\Big]{\frac{1}{m}\sum_{r=1}^m  \diag(\1[r\in \Sc_i]) \Xt_r^\top \Xt_r \otimes a_r(k)a_r(k)^\top } \\
                   &\leq \frac{1}{m}\norm{X}^2\norm{\sum_{r=1}^m a_r(k)a_r(k)^\top  \1[r \in \Sc_i]} \\
                   &\leq \frac{\lambda_n}{m} \norm{A(k)}^2 \\
                   &\leq \frac{4\lambda_n \norm{A(0)}^2}{m},
\end{align*}
where we use the assumption $R_a \leq \norm{A(0)}$. Similarly, using $R_w \le \norm{W(0)} $ we have
\begin{align*}
  \norm{H(k)^\bot} &= \norm[\Big]{\frac{1}{m}\sum_{r=1}^m  \diag(\1[r\in \Sc_i]) \phi(\XT w_r(k)) \phi(w_r(k)^\top X) \otimes I} \\
                   &\leq \frac{1}{m}\norm{X}^2  \norm{W(k)}^2 \\
                   &\leq \frac{4\lambda_n\norm{W(0)}^2}{m}.
\end{align*}
Moreover, using a standard boun on sub-Gaussian matrices, we have $\norm{W(0)} \leq 2\sqrt{m} + \sqrt{d}$ and $\norm{A(0)} \leq 2\sqrt{m} + \sqrt{d}$ with probability at least $1 - 2\exp(-m)$.
Then,
\begin{align*}
  C_2 &= \frac{2\eta}{d} \vec( X - U(k) )^\top K(k)^\bot \vec( X - U (k) )  \\
      &\leq \frac{8 \eta\lambda_n}{d}
\end{align*}
with probability at least $1-2\exp(-m)$.

\qedhere

We re-state the results in the proof in Lemma \ref{lmMovementInTimeWr_joint} and Lemma \ref{lmMovementInTimeAr_joint} to bound the remaining terms $C_3, C_4, C_5$:
\begin{equation}
  \label{eqnBoundGradWr_joint}
    \norm{\nabla_{W}L( W(k), A(k) )}_F \leq \frac{\sqrt{\lambda_n}} {\sqrt{md}} \norm{X - U(k)}_F\norm{A(k)}.
\end{equation}


\begin{equation}
  \label{eqnBoundGradAr_joint}
    \norm{\nabla_{A}L( W(k), A(k) )}_F \leq \frac{\sqrt{\lambda_n} } {\sqrt{md}} \norm{X - U(k)}_F\norm{W(k)}.
\end{equation}

\begin{Claim}
  \label{clmC3_joint}
  Let $C_3 = -\frac{2\eta}{d} \vec( X - U(k) )^\top v_2$. We have
  $$C_3 \leq \frac{16\eta^2}{d} \sqrt{n\lambda_n} \norm{X - U(k)}_F^2$$ with probability at least $1-3\exp(-m)$.
\end{Claim}
\proof
We have $\vec( X - U(k) )^\top v_2 \leq \norm{v_2}\norm{X - U(k)}_F$, so we need to bound $\norm{v_2}$. Let $D_i = \diag(\1[1 \in \Sc_i], \dots, \1[m \in \Sc_i])$, then:
\begin{align*}
  \norm{v_2}^2 &= \sum_{i=1}^n \norm{v_{2, i}}^2 \\
               &= \sum_{i=1}^n \norm[\Big]{\frac{1}{ \sqrt{md} } \sum_{r \in \Sc_i}  \left( a_r(k+1) \phi( w_r(k+1)^\top x_i ) - a_r(k)\phi(w_r(k)^\top x_i ) \right)}^2 \\
               &= \frac{1}{ md}  \sum_{i=1}^n \norm[\Big]{A(k+1)D_i \phi( W(k+1)^\top x_i ) - A(k)D_i\phi(W(k)^\top x_i )}^2 \\
               &\leq \frac{2\eta^2}{ md }  \sum_{i=1}^n \norm[\Big]{(\nabla_AL) D_i \phi( W(k+1)^\top x_i )}^2 + \norm[\Big]{A(k)D_i(\nabla_WL)^\top x_i )}^2 \\
               &\leq \frac{2n\eta^2}{md } \left( \norm{\nabla_AL}_F^2 \norm{W(k+1)}^2 + \norm{A(k)}^2 \norm{\nabla_WL}_F^2 \right) \\
               &\leq   \frac{2n\lambda_n\eta^2}{m^2d^2 } \norm{X - U(k)}_F^2  \left( \norm{W(k+1)}^2\norm{W(k)}^2 + \norm{A(k)}^4 \right) \\
               &\leq   \frac{64n\lambda_n\eta^2}{d^2 } \norm{X - U(k)}_F^2 
\end{align*}
with probability at least $1 - 3\exp(-m)$. Since $\norm{W(k+1)} \leq 2\norm{W(0)}$ and $\norm{A(k+1)} \leq 2\norm{A(0)}$. Therefore, with that probability
\[
  C_3 \leq \frac{16\eta^2}{d} \sqrt{n\lambda_n}\norm{X - U(k)}_F^2.
\]

\qedhere

\begin{Claim}
  \label{clmC4_joint}
  Let $C_4 = -\frac{2\eta}{d} \vec( X - U(k) )^\top v_3$. We have
  $$C_3 \leq \frac{8\eta^2}{d} \sqrt{n\lambda_n^2} \norm{X - U(k)}^2_F \norm{X - U(0)}_F$$ with probability at least $1-2\exp(-m)$.
\end{Claim}
\proof
We have $\vec( X - U(k) )^\top v_3 \leq \norm{v_3}\norm{X - U(k)}_F$. We want to bound $\norm{v_3}$. Let $D'_i = \diag(\1[ w_1(k)^\top x_i \geq 0], \dots, \1[ w_m(k)^\top x_i \geq 0])$
\begin{align*}
  \norm{v_3}^2 &= \sum_{i=1}^n \norm{v_{3, i}}^2 \\
               &= \sum_{i=1}^n \norm[\Big]{\frac{\eta^2}{\sqrt{md}} \sum_{r \in S_i} (\nabla_{a_r} L) (\nabla_{w_r} L)^\top x_i\1[ w_r(k)^\top x_i \geq 0]}^2 \\
               &\leq \frac{\eta^4}{md}  \norm[\Big]{(\nabla_{A} L) ((\nabla_{W} L) D'_i)^\top x_i}^2 \\
               &= \frac{\eta^4n}{md}  \norm{\nabla_{A} L}_F^2 \norm{ \nabla_{W} L }_F^2 \\
               &\leq \frac{\eta^4n\lambda_n^2}{m^2d^2} \norm{X - U(k)}_F^4 \norm{A(k)}^2\norm{W(k)}^2 \\
               &\leq \frac{16\eta^4 n\lambda_n^2 }{d^2} \norm{X - U(k)}_F ^4 
\end{align*}
with probability at least $1-2\exp(-m)$. Therefore,
\begin{align*}
  C_3 \leq \frac{8\eta^2 \sqrt{n\lambda_n^2} }{d} \norm{X - U(k)}_F ^3 \leq \frac{8\eta^2 \sqrt{n\lambda_n^2} }{d} \norm{X - U(k)}_F^2  \norm{X - U(0)}_F 
\end{align*}
since $\norm{X - U(k)}_F \le \norm{X - U(0)}_F$ by the induction hypothesis.

\qedhere

\begin{Claim}
  \label{clmC5_joint}
  Let $C_5  = \norm{ U (k+1) - U(k) }_F^2$. Then we have
  \begin{align*}
    C_5 \leq \frac{64 \eta^2  \lambda^2_n}{md}\norm{X - U(k)}_F^2.
  \end{align*}
  with probability at least $1 - 3\exp(-m)$.
\end{Claim}
\proof
We bound this by re-iterating the proof of Claim \ref{clmC3_joint}:
\begin{align*}
  \norm{ U(k+1) - U(k) }^2_F &= \frac{1}{m^2}\norm{ A(k+1)\phi(W(k+1)^\top X) - A(k) \phi(W(k)^\top X) }^2_F \\
                             &\leq \frac{2\eta^2}{ m^2 } \left(  \norm[\Big]{(\nabla_AL) \phi( W(k+1)^\top X)}^2 + \norm{A(k)}^2\norm[\Big]{(\nabla_WL)^\top X)}^2 \right) \\
                             &\leq \frac{2\eta^2 \norm{X}^2}{m^2} \left( \norm{\nabla_AL}_F^2 \norm{W(k+1)}^2 + \norm{A(k)}^2 \norm{\nabla_WL}_F^2 \right) \\
                             &\leq   \frac{2\eta^2 \lambda^2_n}{m^3d} \norm{X - U(k)}_F^2  \left( \norm{W(k+1)}^2\norm{W(k)}^2 + \norm{A(k)}^4 \right) \\
                             &\leq   \frac{64\eta^2 \lambda_n^2}{md } \norm{X - U(k)}_F^2
\end{align*}
with probability at least $1-3\exp(-m)$.

\qedhere


\end{document}